\documentclass[10pt,twocolumn,letterpaper]{article}

\usepackage[pagenumbers]{wacv} 

\usepackage{graphicx}
\usepackage{amsmath}
\usepackage{amssymb}
\usepackage{booktabs}
\usepackage{tabularx}
\usepackage{multirow}
\usepackage{amssymb}
\usepackage{pifont}
\newcommand{\cmark}{\ding{51}}%
\newcommand{\xmark}{\ding{55}}%
\usepackage[export]{adjustbox}
\usepackage{color, colortbl}
\definecolor{LightCyan}{rgb}{0.88,1,1}
\usepackage[accsupp]{axessibility} 

\usepackage{xurl} 
\usepackage[pagebackref,breaklinks,colorlinks]{hyperref}

\usepackage[capitalize]{cleveref}
\crefname{section}{Sec.}{Secs.}
\Crefname{section}{Section}{Sections}
\Crefname{table}{Table}{Tables}
\crefname{table}{Tab.}{Tabs.}

\def\wacvPaperID{784} 
\def\confName{WACV}
\def\confYear{2025}

\begin{document}

\title{Detecting Wildfires on UAVs with Real-time Segmentation Trained by Larger Teacher Models}

\author{Julius Pesonen$^{1}$
\and 
Teemu Hakala$^1$
\and 
V\"{a}in\"{o} Karjalainen$^{1}$
\and 
Niko Koivum\"{a}ki$^1$
\and
Lauri Markelin$^1$
\and
Anna-Maria Raita-Hakola$^2$
\and
Juha Suomalainen$^1$
\and
Ilkka P\"{o}l\"{o}nen$^2$ 
\and 
Eija Honkavaara$^1$\\ 
\vspace{-0.2cm} \\
$^1$Finnish Geospatial Research Institute,\\ National Land Survey of Finland \\
\and \\
\vspace{-0.2cm} \\
$^2$ Faculty of Information Technology,\\ University of Jyväskylä\\
}
%
\maketitle

\begin{abstract}
Early detection of wildfires is essential to prevent large-scale fires resulting in extensive environmental, structural, and societal damage. Uncrewed aerial vehicles (UAVs) can cover large remote areas effectively with quick deployment requiring minimal infrastructure and equipping them with small cameras and computers enables autonomous real-time detection. In remote areas, however, detection methods are limited to onboard computation due to the lack of high-bandwidth mobile networks. For accurate camera-based localisation, segmentation of the detected smoke is essential but training data for deep learning-based wildfire smoke segmentation is limited. This study shows how small specialised segmentation models can be trained using only bounding box labels, leveraging zero-shot foundation model supervision. The method offers the advantages of needing only fairly easily obtainable bounding box labels and requiring training solely for the smaller student network. The proposed method achieved 63.3\% mIoU on a manually annotated and diverse wildfire dataset. The used model can perform in real-time at $\sim$25 fps with a UAV-carried NVIDIA Jetson Orin NX computer while reliably recognising smoke, as demonstrated at real-world forest burning events. Code is available at: \url{https://gitlab.com/fgi_nls/public/wildfire-real-time-segmentation}

\end{abstract}
\section{Introduction}

\begin{figure}[!ht]
    \centering
    \includegraphics[width=\linewidth]{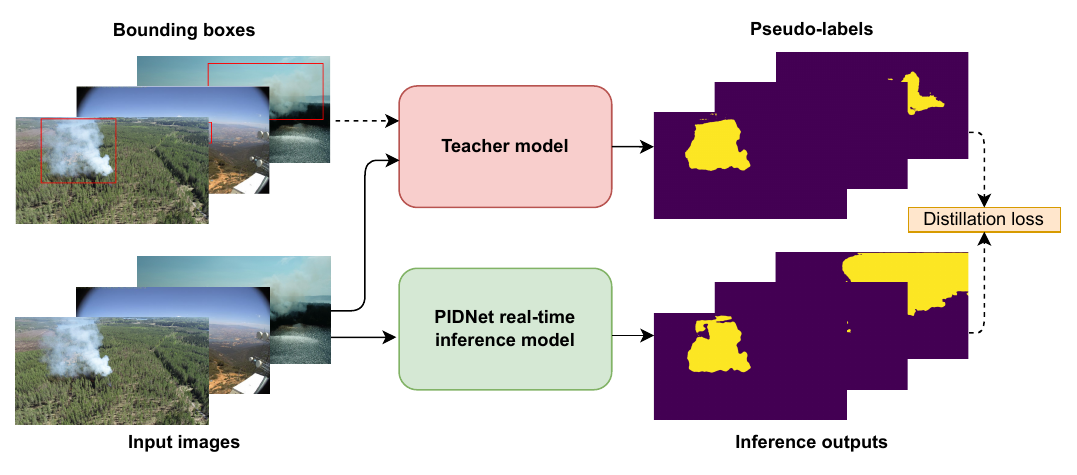}
    \caption{Inference model training scheme. 
    The teacher model is either trained using bounding box supervision method such as BoxSnake~\cite{Yang2023BoxSnakePI} or guided with the bounding boxes like SAM~\cite{kirillov2023segment}.
    }
    \label{fig:scheme}
\end{figure}

The frequency and intensity of extreme wildfires are increasing on Earth, posing tremendous risks to the environment, ecosystems, and societies~\cite{Cunningham2024IncreasingFA}. Early detection is an essential tool to prevent large-scale disasters~\cite{bouguettaya2022review}. Uncrewed aerial vehicle (UAV)-based detection is a promising technology with the potential to enable rapid deployment of surveys over large areas with minimal infrastructure. In remote areas, however, the UAVs are limited to onboard computing for detection due to the lack of high-bandwidth mobile networks. 

Specialised models are required to enable real-time early detection and effective localisation with onboard resources. 
Methods for autonomous detection of wildfires have been studied and the use of UAV-based methods has been proposed as early as 2005~\cite{casbeer2005forest}. Modern deep learning-based computer vision methods, however, have opened new possibilities for obtaining more accurate information more reliably than what has been possible with methods limited to fewer computational resources. 


In this work, we propose using segmentation models capable of accurate real-time detection of wildfires with purely UAV-onboard computing resources. We leveraged bounding box-guided pseudo-label supervision from large foundation models to train the specialised models as simplified in Figure~\ref{fig:scheme}.
We introduce two alternative approaches for supervising the final inference model using either a bounding box-guided zero-shot foundation model teacher or a larger model trained using state-of-the-art bounding box supervision methods.
The methods were evaluated on a diverse human-annotated test set which showed their benefits compared to direct bounding box supervision. Real-world demonstrations and tests on additional data and hardware settings indicated the methods' real-world applicability. The results suggest that largely used bounding box detection methods could be turned into segmentation at minimal costs.



\section{Related Work}

The ideas of UAV-based wildfire monitoring, detection and localisation date back to the early 2000's \cite{casbeer2005forest,merino2006cooperative,casbeer2006cooperative}. However, at the dates of these studies, the available computational resources and non-learning-based computer vision methods posed serious limitations. 
More recent studies have applied bounding box detection to the task of wildfire detection~\cite{Zheng2023ALA,Saydirasulovich2023AnIW,Xiao2023FLYOLOv7AL,Shamta2024DevelopmentOA} and some of the most used bounding box detection models include R-CNN~\cite{Girshick2013RichFH}, Faster R-CNN \cite{Ren2015FasterRT}, YOLO \cite{Redmon2015YouOL}, with its numerous variants, and FPN \cite{Lin2016FeaturePN}. 
Notably, the single-stage object detection networks, including YOLO and its variants focus on the real-time performance of the task and at best can work with very limited computational resources.

The bounding box detections, however, can not fully capture the shape of the detected object which, in the case of UAV-based wildfire detection, could affect the localisation accuracy by even thousands of meters depending on the flight altitude and camera angle. This is key to our study, as we wish to develop a system capable of accurately estimating the 3D location of the fire onboard the UAV. Segmentation combats this issue by capturing the shape of the detected object with pixel-level accuracy. Recent segmentation models, such as ERFNet \cite{Romera2018ERFNetER}, BiSeNet \cite{Yu2018BiSeNetBS}, and ENet \cite{Paszke2016ENetAD} have been proposed specifically for real-time use
and segmentation has also been applied to wildfire detection \cite{Shahid2023ForestFS,Zhang2023FBCANetAS,Wang2023WeaklySF}. However, while the studies have focused on UAV-captured images, their inference times have not been estimated with UAV-carried computers.
Also, the studies mostly focused on flame detection but wildfire smoke is most often visible far further than any flames. The lack of public data, especially such, that is annotated at the pixel level, has largely limited studies on smoke segmentation.

Publicly available wildfire datasets include the FLAME dataset \cite{Shamsoshoara2020AerialIP}, its synthetic extension FLAME Diffuser \cite{wang2024flame} and the Corsican dataset \cite{toulouse2017computer}, all of which focus on flames. The only publicly available smoke focused datasets that could be found at the date of this study were the Bounding Box Annotated Wildfire Smoke Dataset Versions 1.0 and 2.0 by AI For Mankind and HPWREN~\cite{AIforMankind} and a small dataset by the Croatian Center for Wildfire Research~\cite{CroatiaImgs} of 98 images with varying quality pixel-level annotations. 


To tackle the lack of pixel-level annotated wildfire smoke data, we explored weak supervision, which could enable training segmentation models using other than dense pixel-level annotations.
Weakly supervised segmentation refers to training segmentation models using non-pixel-level labels or far fewer labels than the number of input pixels. This can be realised with image-level labels, scribbles, individual pixels, or bounding boxes, as in the case of this study. The earlier solutions for box-supervised training of semantic and instance segmentation have used iterative mask proposals \cite{Dai2015BoxSupEB} or a combination of classical (non-learning-based) and learning-based solutions \cite{Khoreva2016SimpleDI}. 
Many of the recent studies on weakly supervised segmentation have concentrated on instance segmentation which refers to models which differentiate between object instances without any semantic or object class information. Some of the recent proposals include Mask Auto-Labeler (MAL)~\cite{Lan2023VisionTA}, BoxTeacher~\cite{cheng2023boxteacher}, Semantic-aware Instance Mask Generation~\cite{Li2023SIMSI}, and BoxSnake~\cite{Yang2023BoxSnakePI}. For wildfire detection, weakly supervised methods have been proposed to generate synthetic data~\cite{PARK2022103052,ZHENG2024111547}, to produce flame and smoke segmentation from classification models~\cite{9956288}, and to train smoke classification and flame segmentation models~\cite{rs15143606}. While some of those models have been suggested for UAV-collected images, they were not tested for real-time sensing with limited hardware.

To obtain real-time capable segmentation models, we suggest using knowledge distillation.
Knowledge distillation is a way to train effective specialised networks by minimising the difference between the outputs of a larger model, assumed capable of better performance, and those of a smaller, distilled, model~\cite{Hinton2015DistillingTK}. Knowledge distillation has also been applied to semantic segmentation model training for different applications in various forms with positive results~\cite{Liu2019StructuredKD,Qin2021EfficientMI,Yang2022UncertaintyawareCD,Yang2022CrossImageRK}. Knowledge distillation has also been proposed to improve wildfire detection~\cite{song2023fire,xie2023forest} but only for bounding box detection models.

The recent emergence of prompt-guided segmentation~\cite{Lddecke2021ImageSU} and specific models for the task, such as Segment Anything Model (SAM)~\cite{kirillov2023segment} and High-Quality SAM (HQ-SAM) \cite{ke2024segment}, provide new tools for generating pixel-level pseudo-labels from weak labels. The use of combined foundation model and weak label supervision has been shown successful with image level supervision~\cite{Yang2023FoundationMA,Sun2023AnAT,Chen2023SegmentAM,Chen2023WeaklySupervisedSS} alongside Class Activation Maps (CAMs)~\cite{Zhou2015LearningDF} or language understanding from language-image foundation models, such as Grounded-DINO \cite{Liu2023GroundingDM} or CLIP (Contrastive Language-Image Pre-training)~\cite{Radford2021LearningTV}. Foundation models have been applied to synthetic annotated wildfire flame image generation~\cite{wang2024flame} but distilling the information to real-time models has not been studied nor the applicability to UAV-based systems in general.
However, the use of combined bounding box and foundation model supervision for training use case-specific models has been proposed for planetary geological mapping \cite{julka2023knowledge}, cell nuclei segmentation \cite{Cui2023AllinSAMFW}, and remote sensing \cite{Wang2023ScalingupRS}. For cell segmentation of histopathological imagery, such a model has been compared to methods that also use SAM at test time \cite{Tyagi2023GuidedPI}. 
These studies suggest that such method could also be applied to train real-time wildfire smoke detection.

This study fills the gaps in wildfire detection research by showing that real-time detection can be performed using onboard computers with pixel-level accuracy and thus later enable full detection and localisation workflow with purely onboard resources.
Besides only improving UAV-based wildfire detection methods, our work aims to show how numerous resource-limited bounding box detection methods could be practically changed to segmentation.

\section{Methods}

\begin{figure}[!ht]
    \centering
    \includegraphics[height=2.5cm, keepaspectratio]{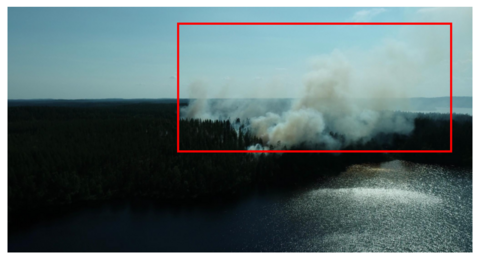}
    \includegraphics[height=2.5cm, keepaspectratio]{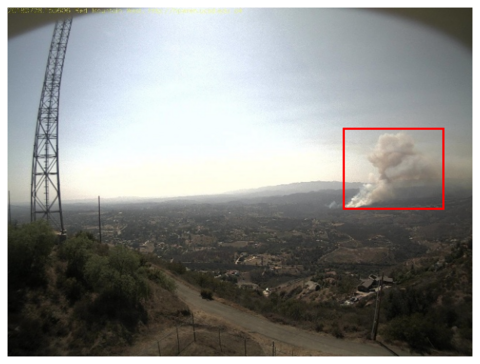}\\
    \includegraphics[height=2.6cm, keepaspectratio]{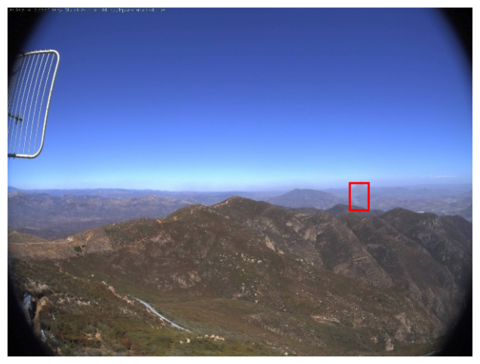}
    \includegraphics[height=2.6cm, keepaspectratio]{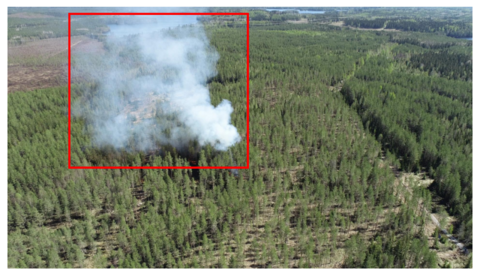}\\
    \caption{Examples from the used bounding box labelled wildfire smoke datasets. The ones with a 4:3 aspect ratio are from the AI For Mankind datasets and the rest are from the UAV dataset.}
    \label{fig:box_samples}
\end{figure}

In this study, we address the gaps in the conducted literature by analysing the performance of PIDNet~\cite{xu2023pidnet} models trained for the segmentation of wildfire smoke using labels obtained from a larger teacher model. We evaluated different teacher models for training the inference models by comparing the resulting accuracy of the final PIDNets. In the proposed framework, the teacher model is a Segment Anything Model guided by the bounding box labels to generate pseudo-labels for supervising the inference model. We compared this method to an alternative, where the teacher model is a larger binary segmentation model trained using state-of-the-art bounding box supervision methods. 

\begin{figure*}[!t]
    \centering
    \includegraphics[width=0.8\textwidth]{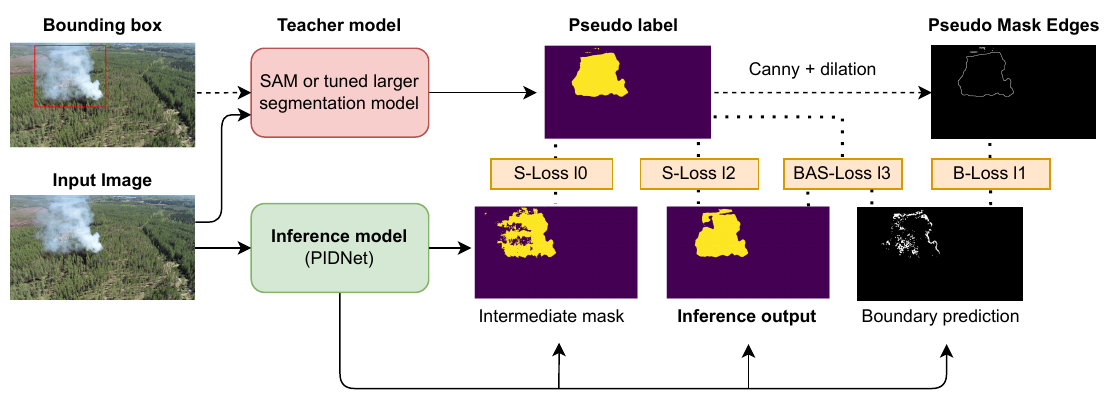}
    \caption{The distillation scheme used to train the final inference model. The label edges are generated with Canny edge detection~\cite{Canny1986ACA} on the label mask image.}
    \label{fig:distillation}
\end{figure*}

We chose PIDNet as the inference architecture due to its capability of real-time performance even with limited computational resources. PIDNet is a three-branch model where each branch has their own focus: local semantic features, global semantic features, and boundary detection. The predictions are combined for the final inference segmentation output. The training method can be formulated as a knowledge distillation scheme with a teacher and a student model, shown in Figure~\ref{fig:distillation}. 


\subsection{Data}
\label{sec:data}

For training and validating the models, we used the Bounding Box Annotated Wildfire Smoke Dataset Versions 1.0 and 2.0 by AI For Mankind and HPWREN, which in this study are referred to as the AI For Mankind data, as well as UAV-collected and manually bounding box labelled data, introduced by Raita-Hakola et al.~\cite{raita2023combining}, which in this study is referred to as the UAV Data. Samples from the datasets are shown in Figure~\ref{fig:box_samples}. The UAV dataset used for this study consists of 5035 bounding box annotated UAV captured aerial footage of boreal forest. 4749 or 94\% of the images contain visible smoke originating from real burning forests. The rest of the images are captured from similar views without any visible wildfire smoke. In addition, another set of 623 smokeless images by AI For Mankind and HPWREN was used to estimate the models' tendency to false positive predictions. To test the models' generalisation to different locations we also computed the metrics on an additional 49-image human-labelled test set by the Croatian Center for Wildfire Research~\cite{CroatiaImgs}. The test set contained two types of segmentation labels, \textit{smoke} and \textit{maybe smoke}, of which the latter was used for the testing due to better representing all the smoke we wished the model to capture. Eight of the test images were discarded due to poor label quality.



Both of the two training datasets had their strengths and weaknesses. The UAV dataset represents the application domain very well in the general environment. However, the smoke is often pictured from very close. In the AI For Mankind data, the seen smoke is often far and small, representing a realistic early detection scenario but the environment is very different from a boreal forest where the real-world application is deployed. Thus, both datasets were represented approximately equally in the training.

To enable using the data for training and evaluation of the segmentation model a three-part datasplit, to training, validation and test sets, was performed. 
An approximately 120-second recording separation interval between used frames was used for the UAV data to avoid major similarities between the sets.
For the AI For Mankind data, the datasplit was performed so that none of the three parts contained images from the same location or direction based on clustering and visual assessment. 

For testing, a total of 80 images, 40 from both the AI For Mankind and UAV datasets, were human-labelled with pixel-wise masks corresponding to the smoke in the image. These were used to compute the final model performance metrics, shown in Section~\ref{results}. Due to the subjective nature of smoke, particularly in the edge regions, and the fairly small number of test samples, small differences in the test metrics should be considered carefully. However, the variety of test scenarios was maximized and the test accuracies can be deemed conclusive between methods with larger differences. The resulting datasplit is shown in Table~\ref{tab:datasplit}. The difference between the total number of UAV data samples in Table~\ref{tab:datasplit} and the earlier presented total number of samples is caused by images discarded due to large similarities.

\begin{table}[!ht]
    \centering
    \begin{tabularx}{\linewidth}{ 
   >{\raggedright\arraybackslash}X| 
   >{\raggedleft\arraybackslash}X
   >{\raggedleft\arraybackslash}X 
   >{\raggedleft\arraybackslash}X 
   >{\raggedleft\arraybackslash}X }
       \textbf{Use case} & \textbf{AI For Mankind} & \textbf{UAV} & \textbf{Total} & \textbf{Portion of the data} \\
       \hline
       Training & 2068 & 1184 & 3252 & 80.6\% \\
       Validation & 453 & 248 & 701 & 17.4\% \\
       Testing & 40 & 40 & 80 & 2.0\% \\
       \hline
       Total & 2561 & 1472 & 4033 & 100\% \\
       \multicolumn{5}{c}{}
    \end{tabularx}
    \caption{The number of used images from both datasets for training, validation and testing.}
    \label{tab:datasplit}
\end{table}

The main evaluation metric for comparing the methods and evaluating the performance of the final model in this study is the mean of the sample-wise Jaccard Index denoted as mIoU (mean intersection over union) and computed as
\begin{equation}
    mIoU(A,B) =  \dfrac{\sum^N_{n=1} \frac{|A_n \cap B_n|}{|A_n \cup B_n|}}{N},
\end{equation}
where $N$ denotes the total number of samples and $A_n$ and $B_n$ denote the predicted and the label masks of the individual samples respectively. In addition, the accuracy, precision, recall, and $F_1$ scores are reported in the main model comparison and their respective equations are shown in the Appendix~\cref{eq:acc,eq:prec,eq:rec,eq:f1}. 

\subsection{Pseudo-labels}

Since the original data was only bounding box labelled, we leveraged larger models to obtain segmentation masks, which were used as pseudo-labels to train the final inference model. The first proposed approach used masks directly generated with SAM using the bounding boxes as guiding prompts, without additional training or fine-tuning of the model. In the second approach, a larger model was trained using different bounding box supervision methods and the pseudo-masks for the whole dataset were obtained by inferring the trained model. 

The Segment Anything Model is a prompt-guided zero-shot segmentation model, which takes two inputs, an image and a point or a bounding box as the prompt. In this study, we used the bounding box prompting to generate masks for each image in the dataset based on the human-annotated bounding boxes. The SAM-generated masks were then used as one of the methods for pseudo-label supervision of the inference model. We used the SAM variant with the ViT-H backbone~\cite{Dosovitskiy2020AnII} to generate the pseudo-labels.

The second approach required training the teacher model using a bounding box supervision method for which we leveraged the state-of-the-art bounding box supervision method BoxSnake~\cite{Yang2023BoxSnakePI}. We adopted the settings introduced for the COCO dataset~\cite{Lin2014MicrosoftCC} while adjusting the number of training iterations to 10\ 000 with a batch size of 16 for the ResNet backbone models, 20\ 000 iterations with a batch size of 8 for the smaller Swin-B backbone model and to 27\ 000 iterations with a batch size of 6 for the Swin-L backbone model. All iteration numbers were adjusted to approximately match the 50 epoch training time of the PIDNet models. The smaller batch sizes were used for the larger models to enable training on the same hardware. 
The different teacher-generated masks were evaluated alongside the student models.


\subsection{Inference model training}

We used AdamW~\cite{loshchilov2017decoupled} with a starting learning rate of 1e-3 and weight decay coefficient of 1e-2 to optimise the inference models. The models were trained for 50 epochs and the validation loss and mean intersection over union was computed on each epoch, from which the best-performing model was chosen for evaluation on the test set and real-world inference based on the validation set mIoU. Minibatches of 16 images were used for the training and the models were pretrained on ImageNet~\cite{Deng2009ImageNetAL}. 

The PIDNets were optimised with four separate loss functions. Two loss functions were used to optimise the two different segmentation outputs and one took into account both the boundary prediction and the final segmentation. In addition, the boundary prediction was optimised with one more separate loss function. In this study, we adapted these four loss functions for the binary smoke segmentation task to distil the knowledge from the larger teacher model. The inference output, which is the final segmentation output of the model, was used alone for evaluating the model. The full distillation scheme is shown in Figure~\ref{fig:distillation}.

The total distillation loss is computed as
\begin{equation}
    Loss = \lambda_0 l_0 + \lambda_1 l_1 + \lambda_2 l_2 + \lambda_3 l_3,
\end{equation}
where loss functions $l_0$ and $l_1$ are binary cross entropy losses from the two alternative segmentation outputs of the PIDNet. $l_2$ is a weighted boundary-awareness cross entropy loss~\cite{Takikawa2019GatedSCNNGS} and $l_3$ is a BAS-Loss that is a binary cross entropy loss computed only at the pixels where the boundary output of the model exceeds the threshold $t$.
For the weighting of the losses we used the parameters $\lambda_0 = 0.4$, $\lambda_1 = 20$, $\lambda_2 = 1$, $\lambda_3 = 1$ and $t = 0.8$ as set empirically in the study introducing the PIDNet model.

The training samples were augmented randomly, similar to TrivialAugment \cite{Muller_2021_ICCV} where a single augmentation with a random strength was randomly chosen for a sample at training time from a predefined set of augmentations. We opted for the fairly heavy augmentation strategy to overcome the limitations of the small amount of data and possibly noisy pseudo-labels. The random augmentation space included cropping, vertical flips, rotation, perspective, erasing, grayscale, gaussian blur, inversion, sharpness, and colour jitter. All of them were applied with the same probability and, in addition, a horizontal flip was applied 50\% of the time regardless of other augments.

\subsection{UAV-based system and testing}
\label{sec:prac_sys}

A UAV-carried sensor and computation system were implemented to test the performance of the obtained model in a real-world scenario. The lightweight system consisted of a small global shutter RGB camera AR0234~\cite{arducamAR0234Arducam}, with a 1920$\times$1080~px resolution and 90° horizontal and 75° vertical field of view, and an NVIDIA Jetson Orin NX~\cite{jetson} computer. The total system weight was only $\sim$200 grams meaning that small UAVs could easily carry it. The system was tested at real-world forest-burning events to demonstrate the capability of the real-time segmentation model. 

The burning events were organised by the Finnish Mets\"{a}hallitus (a governmental forest administration organisation) at conservation areas in Southern Finland. The tests were conducted by recording the model outputs and the corresponding inputs as videos on the UAV-carried system from varying distances between one and fifteen kilometres. The sensor system was carried by a DJI Matrice 300 RTK UAV~\cite{djiSpecsMATRICE}, flown at an altitude of 120 meters above ground level. The test flights were conducted during sunny weather in June. To ensure that video was captured where the smoke was in the view of the camera, the small AR0234 camera was mounted on a Zenmuse H20T~\cite{djiSpecsZenmuse}, with a continuous video stream to the UAV pilot.

\begin{table*}
    \centering
    \begin{tabularx}{\textwidth}{ 
   >{\raggedright\arraybackslash}p{4cm}
   >{\raggedright\arraybackslash}p{2cm} 
   >{\raggedleft\arraybackslash}X
   >{\raggedleft\arraybackslash}X
   >{\raggedleft\arraybackslash}X
   >{\raggedleft\arraybackslash}X
    >{\raggedleft\arraybackslash}X
   }
        \textbf{Teacher} & \textbf{Student} & \textbf{mIoU} & \textbf{Accuracy} & \textbf{Precision} & \textbf{Recall}  &  $\boldsymbol{F_1}$ \\
        \hline
        ResNet-50-RCNN-FPN & - & 0.560 & 0.959 & 0.748 & 0.669 & 0.674 \\
        ResNet-101-RCNN-FPN & - & 0.573 & 0.964 & 0.725 & 0.684 & 0.686 \\
        Swin-B-FPN & - & \underline{0.641} & \textbf{0.967} & 0.860 & \underline{0.720} & \underline{0.760} \\
        Swin-L-FPN & - & \textbf{0.651} & \underline{0.966} & \underline{0.871} & \textbf{0.730} & \textbf{0.772} \\
        SAM* & - & 0.636 & 0.958 & \textbf{0.912} & 0.693 & 0.754 \\
        
        \rowcolor{LightCyan}
        ResNet-50-RCNN-FPN & PIDNet-S & \textbf{0.602} & 0.962 & 0.800 & \textbf{0.727} & \textbf{0.729} \\
        \rowcolor{LightCyan}
        ResNet-101-RCNN-FPN & PIDNet-S & 0.571 & 0.957 & \underline{0.881} & 0.626 & 0.689 \\
        \rowcolor{LightCyan}
        Swin-B-FPN & PIDNet-S & 0.582 & \underline{0.964} & 0.797 & 0.665 & 0.696 \\
        \rowcolor{LightCyan}
        Swin-L-FPN & PIDNet-S & 0.591 & \textbf{0.967} & 0.820 & \underline{0.681} & \underline{0.710} \\
        \rowcolor{LightCyan}
        SAM & PIDNet-S & \underline{0.594} & 0.960 &\textbf{0.891} & 0.645 & 0.707  \\
        
        ResNet-50-RCNN-FPN & PIDNet-M & \underline{0.636} & 0.966 & 0.824 & \textbf{0.741} & \underline{0.756} \\
        ResNet-101-RCNN-FPN & PIDNet-M & 0.597 & 0.963 & \underline{0.880} & 0.633 & 0.710 \\
        Swin-B-FPN & PIDNet-M & \textbf{0.643} & \underline{0.967} & 0.838 & \underline{0.737} & \textbf{0.761} \\
        Swin-L-FPN & PIDNet-M & 0.629 & \textbf{0.968} & 0.856 & 0.710 & 0.743 \\
        SAM & PIDNet-M & 0.606 & 0.960 & \textbf{0.884} & 0.666 & 0.721 \\
        
        \rowcolor{LightCyan}
        ResNet-50-RCNN-FPN & PIDNet-L & \underline{0.613} & \textbf{0.966} & \underline{0.841} & \underline{0.711} & \underline{0.732} \\
        \rowcolor{LightCyan}
        ResNet-101-RCNN-FPN & PIDNet-L & 0.611 & 0.962 & \textbf{0.865} & 0.654 & 0.721 \\
        \rowcolor{LightCyan}
        Swin-B-FPN & PIDNet-L & \textbf{0.618} & \textbf{0.966} & 0.825 & \textbf{0.720} & \textbf{0.736} \\
        \rowcolor{LightCyan}
        Swin-L-FPN & PIDNet-L & 0.591 & \textbf{0.966} & 0.768 & 0.700 & 0.706 \\
        \rowcolor{LightCyan}
        SAM & PIDNet-L & 0.594 & 0.956 & 0.809 & 0.699 & 0.711 \\
    \end{tabularx}
    \caption{Test set metrics of the different models. In all metrics higher is better. All teacher models besides SAM were trained using BoxSnake. *The SAM test metrics are computed from the bounding box guided masks. The best results for each student model are in bold and the second best are underlined.
    }
    \label{tab:models}
\end{table*}

\section{Results}
\label{results}

Table~\ref{tab:models} shows the model test set metrics obtained using different teacher models. 
Each PIDNet model was trained using the distillation scheme based on the teacher model generated pseudo-labels, shown in Figure~\ref{fig:distillation}. The main evaluation metric is the mIoU, but the accuracy, precision, recall, and F1 scores are displayed for improved analysis. The tested teacher models are the Segment Anything model guided with the bounding boxes and Mask R-CNNs~\cite{He2017MaskR} using feature maps extracted with a feature pyramid network (FPN)~\cite{Lin2016FeaturePN} from different backbones which were ResNet~\cite{He2015DeepRL} and Swin Transformer~\cite{Liu2021SwinTH} variants, trained using the BoxSnake bounding box supervision method. The BoxSnake-trained models are shortened as [Backbone]-RCNN-FPN in Tables~\ref{tab:models} and~\ref{tab:split_res}.

The results show that the distillation scheme successfully transfers the smoke segmentation knowledge from the teacher model to the student. The mIoU obtained with any of the PIDNet variants was consistently better than what was achieved using the BoxSnake-trained ResNet-50-RCNN-FPN, the smallest model trained using direct box supervision. Surprisingly, with smaller teacher networks the distillation training improves the results over the accuracy of the teacher network, most clearly shown in the comparison between the BoxSnake supervised ResNet-50-RCNN-FPN model and the corresponding PIDNet student models. 

Out of the PIDNet variants, PIDNet-M was found to be the best for the task regarding the test set metrics regardless of the teacher model. However, for practice, the PIDNet-S was chosen because of the far superior inference speed on the NVIDIA Jetson Orin NX, shown in Table~\ref{tab:n_params} and the more extensive model speed benchmark with hardware setting variation in the Appendix Table~\ref{tab:hw_bench}.
We expect the consistently worse performance of the PIDNet-L model to be caused by overfitting to noisy pseudo-labels, caused by the teacher models poorly capturing the human perception of the smoke in some images. Still, it is interesting how this behaviour was reflected in all teacher variants.


\begin{table}[!h]
    \centering
    \begin{tabularx}{\linewidth}{l
    >{\raggedleft\arraybackslash}X
    >{\raggedleft\arraybackslash}X
    >{\raggedleft\arraybackslash}X}
        \textbf{Model} & \textbf{N. params} & \textbf{GFLOPs} & \textbf{FPS} \\
        \hline
        PIDNet-S & 7.72M & 49.7 & 25.88 \\
        PIDNet-M & 28.8M & 182.9 & 9.86 \\
        PIDNet-L & 37.3M & 284.3 & 7.47 \\
        \multicolumn{4}{c}{} \\
    \end{tabularx}
    \caption{Model sizes and frame rates. Tests were done with RGB inputs of size 1080 $\times$ 1920 pixels and the inference times were computed on the GPU of NVIDIA Jeston Orin NX using NVIDIA TensorRT~\cite{TensorRT}.}
    \label{tab:n_params}
\end{table}

\begin{table}[!h]
    \centering
    \begin{tabularx}{\linewidth}{l|rr}
        & \multicolumn{2}{c}{\textbf{PIDNet-S mIoU}} \\
       \textbf{Teacher method} & \textbf{UAV} & \textbf{AI For Mank.} \\
       \hline
       BoxSnake R50-RCNN-FPN & 0.718 & \underline{0.486} \\
       BoxSnake R101-RCNN-FPN & 0.671 & 0.470 \\
       BoxSnake Swin-B-FPN & \underline{0.740} & 0.424 \\
       BoxSnake Swin-L-FPN & \textbf{0.753} & 0.429 \\
       SAM & 0.689 & \textbf{0.498} \\
       \multicolumn{3}{c}{} \\
    \end{tabularx}
    \caption{Test errors of the different PIDNet-S methods on the two different datasets. All models were trained on a mixed dataset.}
    \label{tab:split_res}
\end{table}

\begin{table*}[!ht]
    \centering
    \begin{tabular}{>{\centering}p{0.1\textwidth}
    >{\centering}p{0.1\textwidth}
    >{\centering}p{0.1\textwidth}
    >{\centering}p{0.1\textwidth}
    >{\centering}p{0.1\textwidth}
    >{\centering}p{0.1\textwidth}
    >{\centering}p{0.1\textwidth}
    p{0.1\textwidth}}
       \textbf{Input} & \textbf{GT} & \textbf{SAM} & \textbf{PIDNet-S} & \textbf{Input} & \textbf{GT} & \textbf{SAM} & \hspace{2mm}\textbf{PIDNet-S}  \\
       \includegraphics[width=0.119\textwidth]{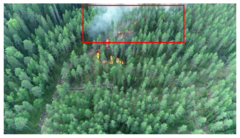}& 
       \includegraphics[width=0.119\textwidth]{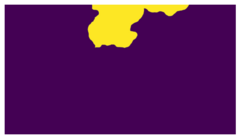}&
       \includegraphics[width=0.119\textwidth]{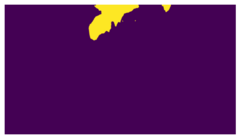} &
       \includegraphics[width=0.119\textwidth]{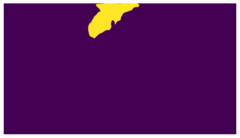} & 
       \includegraphics[width=0.119\textwidth]{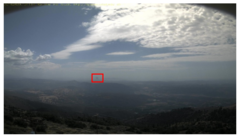}  & 
       \includegraphics[width=0.119\textwidth]{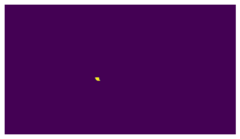}  & 
       \includegraphics[width=0.119\textwidth]{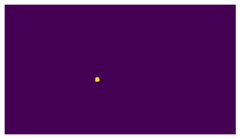}  &  
       \includegraphics[width=0.119\textwidth]{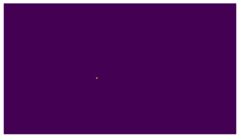}
       \\
       \includegraphics[width=0.12\textwidth]{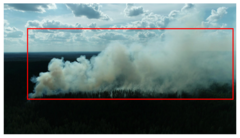}  & 
       \includegraphics[width=0.12\textwidth]{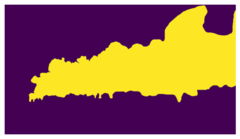}&
       \includegraphics[width=0.12\textwidth]{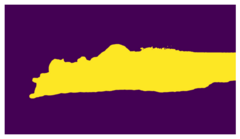}&
       \includegraphics[width=0.12\textwidth]{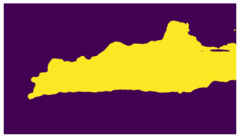}& 
       \includegraphics[width=0.12\textwidth]{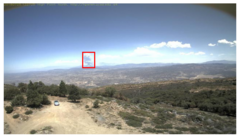} & 
       \includegraphics[width=0.12\textwidth]{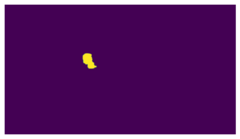} & 
       \includegraphics[width=0.12\textwidth]{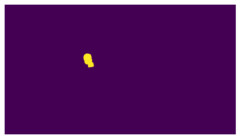} & 
       \includegraphics[width=0.12\textwidth]{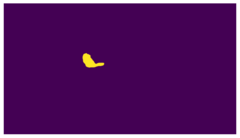}\\
       \includegraphics[width=0.12\textwidth]{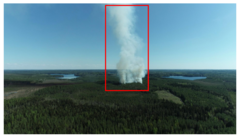}&  
      \includegraphics[width=0.12\textwidth]{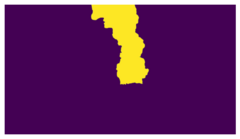} &
       \includegraphics[width=0.12\textwidth]{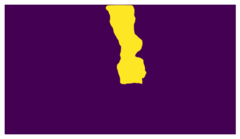}&
       \includegraphics[width=0.12\textwidth]{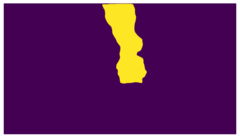}& 
       \includegraphics[width=0.12\textwidth]{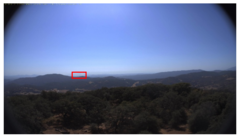}  &
       \includegraphics[width=0.12\textwidth]{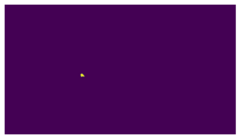} & 
       \includegraphics[width=0.12\textwidth]{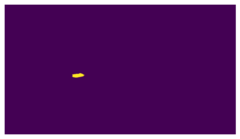}& 
       \includegraphics[width=0.12\textwidth]{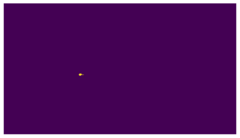} \\
       \includegraphics[width=0.12\textwidth]{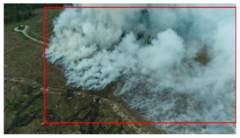} & 
       \includegraphics[width=0.12\textwidth]{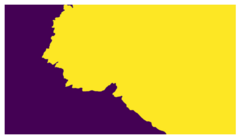}&
       \includegraphics[width=0.12\textwidth]{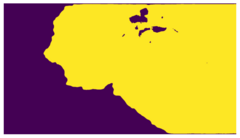} & 
       \includegraphics[width=0.12\textwidth]{imgs/qual_results_compressed/pid_s/ruokolahti_DJI_0087_frame134pidnet_mask2.png}&
       \includegraphics[width=0.12\textwidth]{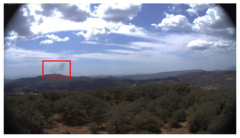}  & 
       \includegraphics[width=0.12\textwidth]{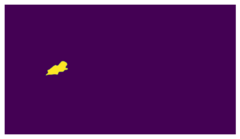}& 
       \includegraphics[width=0.12\textwidth]{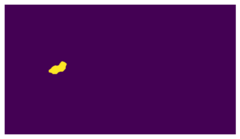}& 
       \includegraphics[width=0.12\textwidth]{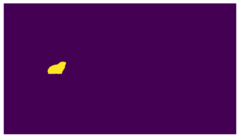}\\
    \end{tabular}
    \caption{Random examples from the manually annotated test set. From left to right in two columns: Input and a manually created bounding box, manual ground truth mask (GT), SAM-generated pseudo label, and the corresponding PIDNet-S prediction.}
    \label{tab:qualitative}
\end{table*}

The lack of correlation between the teacher model metrics and those of the student models on the test set was surprising. The ResNet-50-RCNN-FPN supervised PIDNet models performed well in multiple metrics despite the worse results of their teacher model. The results also differed slightly between the two datasets. The separate mIoU achieved by the PIDNet-S variants on the test samples of the two original datasets are presented in Table~\ref{tab:split_res}. The results show that depending on the teacher the student model performance shifts between the datasets. 


The results of the generalisation tests on the Croatian Center for Wildfire Research data are shown in Table~\ref{tab:croatia_res}. Full student model results are found in  Table~\ref{tab:croatia_fullset} and qualitative results in Table~\ref{tab:croatia_qual} of the supplementary material. The tests were performed without any retraining of the models on the associated training data. While the results were worse than those of the main datasets, they still support the conclusion that the obtained models can be used for general smoke segmentation. It must also be noted that many of the ground truth labels were of questionable quality.

\begin{table}[!h]
    \centering
    \begin{tabularx}{\linewidth}{l|rrrrr}
       \textbf{Model} & \hspace{2mm} \textbf{mIoU} & \textbf{Acc.} & \textbf{Prec.} & \textbf{Rec.} & $\boldsymbol{F_1}$ \\
       \hline
       PIDNet-S  & \underline{0.410} & \underline{0.848} & \underline{0.824} & \underline{0.481} & \underline{0.534} \\
       PIDNet-M  & 0.374 & 0.832 & \textbf{0.867} & 0.406 & 0.502 \\
       PIDNet-L  & \textbf{0.479} & \textbf{0.873} & 0.823 & \textbf{0.552} & \textbf{0.614} \\
    \end{tabularx}
    \caption{Evaluation metrics on the Croatian Center for Wildfire Research data using the different SAM supervised models.}
    \label{tab:croatia_res}
\end{table}

Additional data consisting of images without smoke was also used to investigate the models' tendency for false positives. The false positive rates for each of the SAM-supervised PIDNet models are shown in Table~\ref{tab:false_positives}. The tests revealed that the models produced a significant amount of false positives which should be considered in future studies and practical applications. Likely, a more consistent training data distribution would relieve the problem significantly. 

\begin{table}[!h]
    \centering
    \begin{tabularx}{\linewidth}{l|rrr}
       \textbf{Model} & \hspace{4mm} PIDNet-S  & \hspace{4mm} PIDNet-M & \hspace{4mm} PIDNet-L \\
       \hline
       \textbf{FP rate} & \underline{0.143} & \textbf{0.136} & 0.146 \\
    \end{tabularx}
    \caption{False positive (FP) rates for the SAM supervised models on the AI For Mankind Classification dataset's smokeless images.}
    \label{tab:false_positives}
\end{table}

\subsection{Qualitative results}

As the PIDNet-S model trained using SAM supervision was chosen for the real-world application, qualitative results of SAM and the corresponding PIDNet-S are shown in Table~\ref{tab:qualitative} alongside the input image, the bounding box label, and the human-drawn ground truth mask. For the rest of the models, qualitative results are provided in Appendix \cref{tab:resnet50,tab:resnet101,tab:swin_b,tab:swin_l,tab:sam}.

Besides the individual sample in which the PIDNet model can not distinguish the smoke, any large conclusions were difficult to draw from the qualitative results besides the fact that both the teacher and the student model could segment smoke from the test images of both datasets. However, some variety could be observed in predictions between all trained models, such as more false positives in the PIDNet-L outputs.

\subsection{Ablation}
\label{sec:abl}

We performed an additional ablation on the four different loss terms used for the distillation optimisation to observe their effect on the model inference. The ablation was performed using the SAM-supervised PIDNet-S model by omitting individual loss terms. The results of the loss function ablation are shown in Table~\ref{tab:loss_abl}.

\begin{table}[!h]
    \centering
    \begin{tabularx}{\linewidth}{llll|>{\raggedleft\arraybackslash}X>{\raggedleft\arraybackslash}X}
         \multicolumn{4}{c}{\textbf{Loss functions}} & \multicolumn{2}{c}{\textbf{mIoU}} \\
       \textbf{Seg. $l_0$} & \textbf{Seg. $l_1$} & \textbf{B. $l_2$} & \textbf{BAS $l_3$} & \textbf{Val.} & \textbf{Test}\\
       \hline
       \xmark  & \cmark & \cmark & \cmark & 0.529 & 0.533 \\
       \cmark & \xmark  & \cmark & \cmark & \textbf{0.550} & 0.585 \\
       \cmark & \cmark  & \xmark & \cmark & 0.535 & \textbf{0.633} \\
       \cmark & \cmark  & \cmark & \xmark & 0.533 & \underline{0.608} \\
       \cmark & \cmark & \xmark & \xmark & 0.526 & 0.594 \\
       \hline
       \cmark & \cmark  & \cmark & \cmark & \underline{0.540} & 0.594 
    \end{tabularx}
    \caption{Loss function ablation results.}
    \label{tab:loss_abl}
\end{table}

While the test errors improved notably when one of the loss terms $l_2$ or $l_3$, which depend on the boundary prediction, was omitted, the effect was assumed to be at least partially caused by the supervising model. As the boundaries of the SAM-generated masks were visibly misplaced in multiple situations, weighing the boundary losses too highly could understandably weaken the segmentation performance of the model. The negative effects of the boundary loss terms could also appear due to the changing and unclear nature of smoke in general. However, as the usefulness of these loss terms has been indicated clearly in the studies they've been presented in, omitting their use completely in the teacher model comparison of Table~\ref{tab:models} was deemed unnecessary. In addition, omitting both boundary-dependent loss terms completely yields worse results, so it can be assumed that the optimal setting lies somewhere between the original loss weights and zero. This optimisation was left for future work.

\subsection{Real-world performance}

The model was tested in practice at real forest-burning events described in Section~\ref{sec:prac_sys}. The model ran on the NVIDIA Jetson Orin NX computer carried by the UAV at approximately 6~fps due to the data recording and some CPU preprocessing which could be optimised further. The model itself was capable of inference at $\sim$25~fps, which can be expected for a practical application in which video data is not saved, but the model output is only checked for positives, and an alert is sent in case fire is detected.

\begin{figure}[!ht]
    \includegraphics[width=0.32\linewidth]{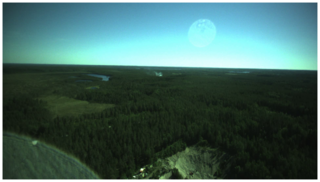} \includegraphics[width=0.32\linewidth]{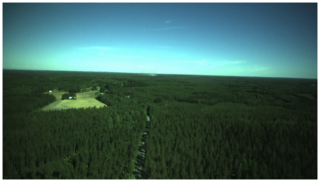} \includegraphics[width=0.32\linewidth]{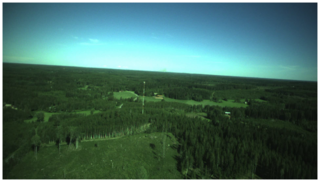} 
    \includegraphics[width=0.32\linewidth]{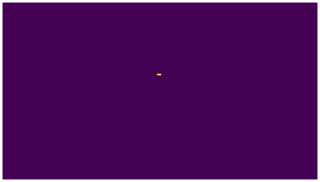} \hspace{0.3px} \includegraphics[width=0.32\linewidth]{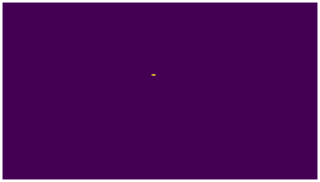} \hspace{0.3px} \includegraphics[width=0.32\linewidth]{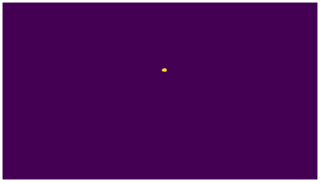} 
    \caption{Visualisations of the practical real-time detection tests. The shown results, from left to right, were achieved from 1.4, 4.0 and 9.7 kilometres away from the burning, in ground distance.}
    \label{fig:rw_inf}
\end{figure}

The model successfully detected the smoke at distances between 1.4 and 9.7~km, even though the presented smoke was small, especially compared to the UAV data used for training. Both inputs and outputs of the model were recorded as videos, some samples of which are provided as supplementary material. The resulting videos showed that, while some false positives appeared in regions without smoke, they were temporally far less consistent than the true positives caused by smoke, meaning they could be filtered with simple temporal constraints. Figure~\ref{fig:rw_inf} shows still frames from the test flight recordings. 


The real-world tests showed that the proposed methodology is suitable for practical applications. Importantly, it also supported the results of the model generalisation, as the test situation was very different from what was seen in any of the training data. The quality of the input images was also far worse in terms of lighting and contrast compared to the UAV training and test data.




\section{Conclusions}

The study shows that smoke segmentation for real-time early detection of wildfires is feasible with computational resources limited to UAV-carried edge computing systems. The models can be trained using bounding box-guided foundation models for pseudo-label supervision. The capability was tested on a diverse offline dataset against manual annotations where the proposed model achieved an accuracy of 63.3\% mIoU. The system was also tested in real-world scenarios where qualitative video results proved the system's detection capability. The inference could run at $\sim$25~fps on the lightweight NVIDIA Jetson Orin NX computer when sufficient results for reliable real-world onboard sensing were achieved with the system running at $\sim$6~fps with the workflow including both the inference and video recording. The model successfully captured smoke from up to 9.7 kilometres in a scenario with little visible smoke. In addition, the generalisation was demonstrated on an additional human-annotated test set from very different geographical locations and viewing angles, where acceptable results were achieved. 

The method is a major leap towards a fully autonomous UAV-based early detection system where the full detection and localisation workflow is performed with the onboard computer alone. Besides improving UAV-based wildfire early detection methods, this study demonstrates how computation-limited bounding box detection methods could be changed to segmentation without additional manual labour or sacrificing the real-time capabilities.


While the generalisation was demonstrated, the greatest limitation was still the narrow geographical and visual distribution of the data. 
Thus, potential future work includes extending the data distribution with other types of weak supervision, such as image-level labelled or diffusion model-generated data \cite{Wu2023DiffuMaskSI,Karazija2023DiffusionMF}. Another limitation was demonstrated by the models' tendency to generate false positive predictions, which we suggest combatting in future studies by expanding the training data with additional smokeless images, possibly including features such as clouds or fog which could be mistaken as smoke. Due to the lack of quality densely labelled data, the study also did not include a comparison to a fully supervised method. We hope to extend the study later with such a comparison. The loss terms should also be optimised further, as shown by the ablation in Section~\ref{sec:abl}.

The proposed method also heavily relies on the quality of the SAM-generated pseudo-labels, which as indicated by the results, were not perfect. Thus, it would be of high value to further explore how the pseudo-label masks could be improved. Some simple strategies, such as adding variation to the prompts by random shifting of the manually generated bounding boxes or by assigning randomly chosen points inside the box as prompts could prove beneficial. The variationally generated masks could be used either as label augmentations or to produce more consistent pseudo-labels. Another possibility would be to fine-tune SAM specifically for the smoke segmentation task.

\section*{Acknowledgments}
This research was funded by the Academy of Finland within project Fireman (decision no. 346710, 348009) and ML4DRONE (decision no. 357380). The FireMan project is funded under the EU’s Recovery and Resilience Facility that promotes the green and digital transitions through research. This study has been performed with affiliation to the Academy of Finland Flagship Forest–Human–Machine Interplay—Building Resilience, Redefining Value Networks and Enabling Meaningful Experiences (UNITE) (decision no. 357908).

{\small
\bibliographystyle{ieee_fullname}
\bibliography{main}
}

\clearpage
\section*{Appendix}

The additional binary metrics used in Table~\ref{tab:models}:
\begin{equation}
    Accuracy = \dfrac{TP + TN}{P + N}
    \label{eq:acc}
\end{equation}

\begin{equation}
    Precision = \dfrac{TP}{TP + FP}
    \label{eq:prec}
\end{equation}

\begin{equation}
    Recall = \dfrac{TP}{TP + FN}
    \label{eq:rec}
\end{equation}

\begin{equation}
    F_1 = \dfrac{2 TP}{2 TP + FP + FN},
    \label{eq:f1}
\end{equation}
where $P$ and $N$ denote positive and negative ground truth values and $TP,TF,FP$, and $FN$ denote the corresponding true and false predictions. 

\begin{table*}[!t]
    \centering
    \begin{tabular}{>{\centering}p{0.135\textwidth}
    >{\centering}p{0.135\textwidth}
    >{\centering}p{0.135\textwidth}
    >{\centering}p{0.135\textwidth}
    >{\centering}p{0.135\textwidth}
    p{0.135\textwidth}}
       \textbf{Input} & \textbf{GT} & \textbf{ResNet-50} & \textbf{PIDNet-S} & \textbf{PIDNet-M} & \hspace{6mm}\textbf{PIDNet-L}  \\
       \includegraphics[width=0.155\textwidth]{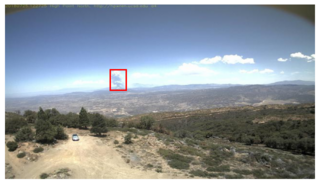} & 
        \includegraphics[width=0.155\textwidth]{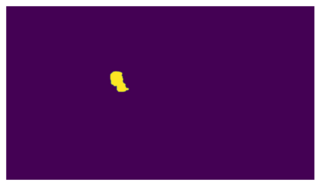} & 
        \includegraphics[width=0.155\textwidth]{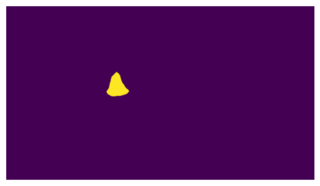} &
        \includegraphics[width=0.155\textwidth]{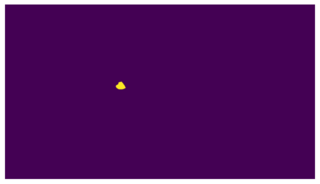} &
        \includegraphics[width=0.155\textwidth]{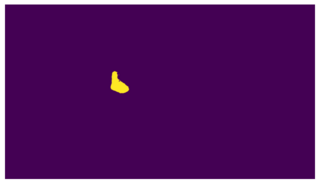} &
        \includegraphics[width=0.155\textwidth]{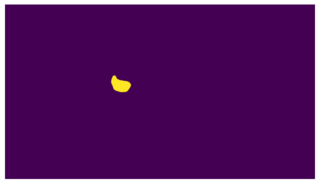} \\

        \includegraphics[width=0.155\textwidth]{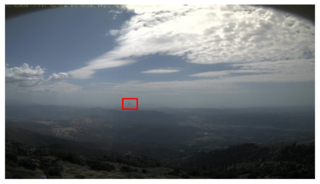} & 
        \includegraphics[width=0.155\textwidth]{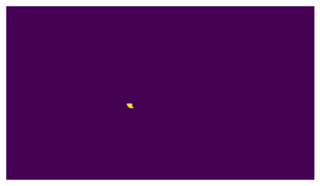} & 
        \includegraphics[width=0.155\textwidth]{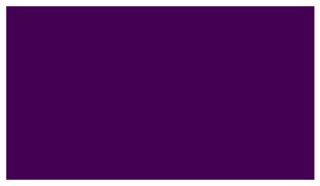} &
        \includegraphics[width=0.155\textwidth]{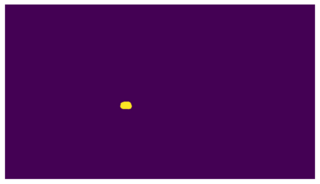} &
        \includegraphics[width=0.155\textwidth]{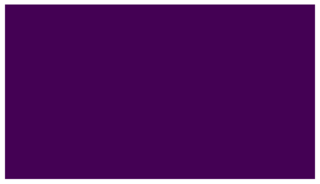} &
        \includegraphics[width=0.155\textwidth]{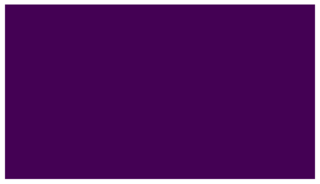} \\

        \includegraphics[width=0.155\textwidth]{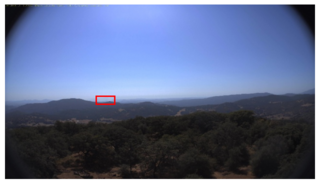} & 
        \includegraphics[width=0.155\textwidth]{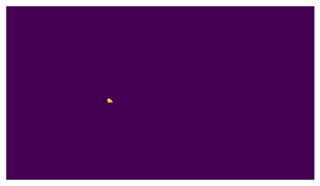} & 
        \includegraphics[width=0.155\textwidth]{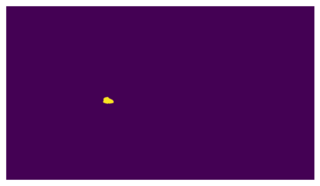} &
        \includegraphics[width=0.155\textwidth]{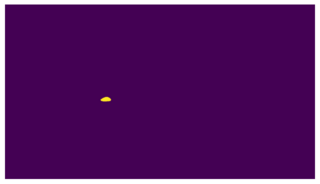} &
        \includegraphics[width=0.155\textwidth]{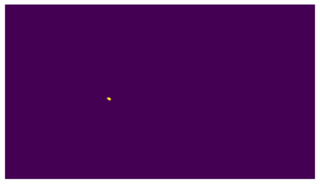} &
        \includegraphics[width=0.155\textwidth]{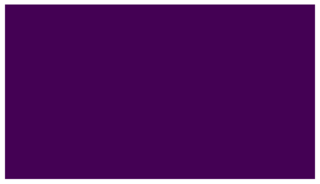} \\

        \includegraphics[width=0.155\textwidth]{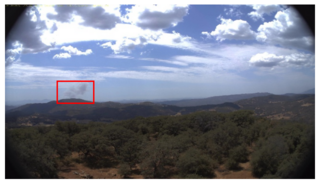} & 
        \includegraphics[width=0.155\textwidth]{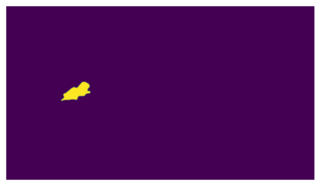} & 
        \includegraphics[width=0.155\textwidth]{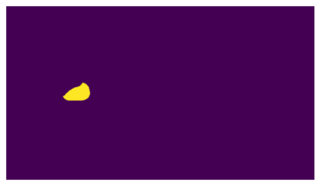} &
        \includegraphics[width=0.155\textwidth]{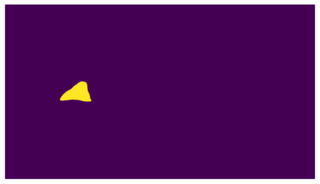} &
        \includegraphics[width=0.155\textwidth]{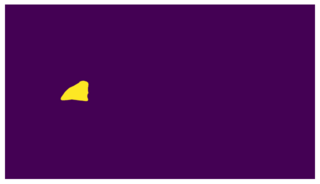} &
        \includegraphics[width=0.155\textwidth]{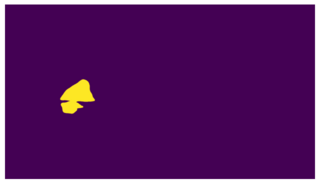} \\

        \includegraphics[width=0.155\textwidth]{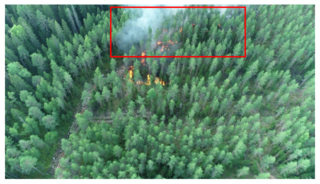} & 
        \includegraphics[width=0.155\textwidth]{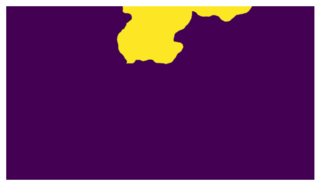} & 
        \includegraphics[width=0.155\textwidth]{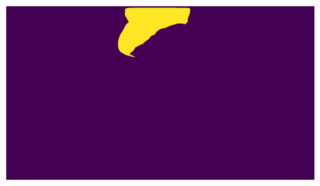} &
        \includegraphics[width=0.155\textwidth]{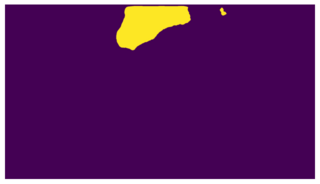} &
        \includegraphics[width=0.155\textwidth]{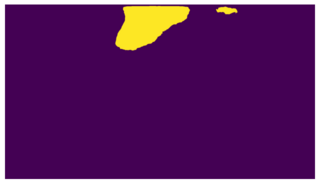} &
        \includegraphics[width=0.155\textwidth]{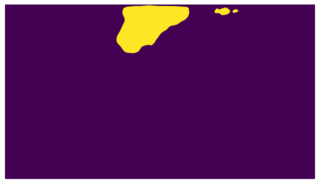} \\

        \includegraphics[width=0.155\textwidth]{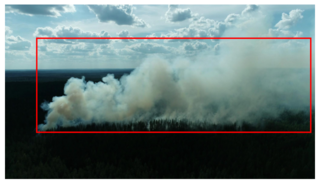} & 
        \includegraphics[width=0.155\textwidth]{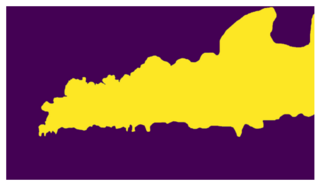} & 
        \includegraphics[width=0.155\textwidth]{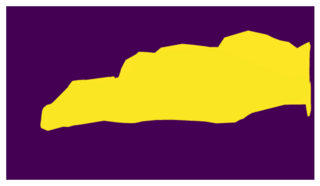} &
        \includegraphics[width=0.155\textwidth]{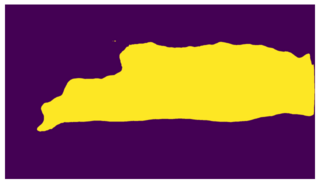} &
        \includegraphics[width=0.155\textwidth]{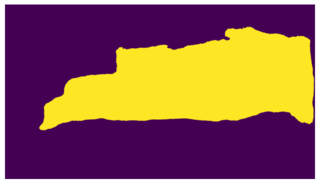} &
        \includegraphics[width=0.155\textwidth]{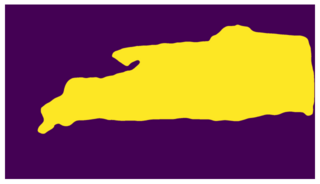} \\

        \includegraphics[width=0.155\textwidth]{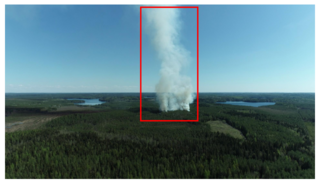} & 
        \includegraphics[width=0.155\textwidth]{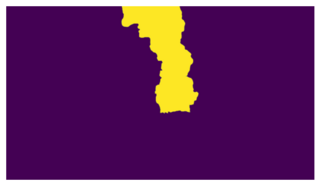} & 
        \includegraphics[width=0.155\textwidth]{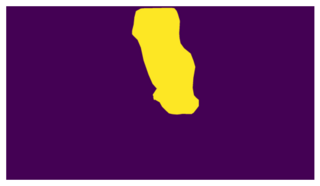} &
        \includegraphics[width=0.155\textwidth]{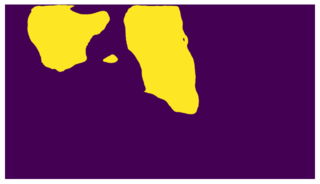} &
        \includegraphics[width=0.155\textwidth]{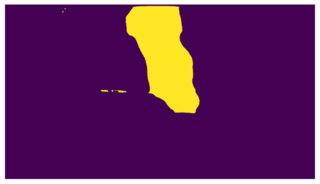} &
        \includegraphics[width=0.155\textwidth]{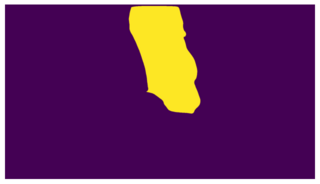} \\

        \includegraphics[width=0.155\textwidth]{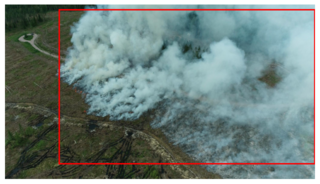} & 
        \includegraphics[width=0.155\textwidth]{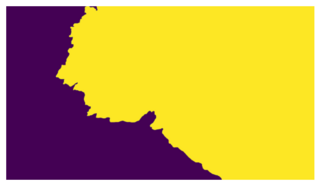} & 
        \includegraphics[width=0.155\textwidth]{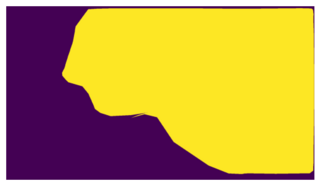} &
        \includegraphics[width=0.155\textwidth]{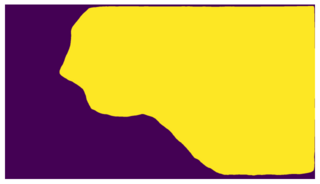} &
        \includegraphics[width=0.155\textwidth]{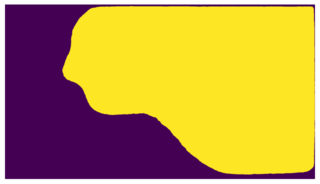} &
        \includegraphics[width=0.155\textwidth]{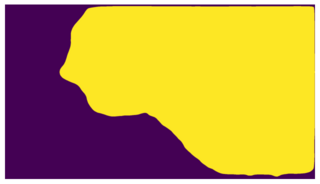} \\
    \end{tabular}
    \caption{Visualisation of outputs obtained using the BoxSnake supervised ResNet-50 and the corresponding distilled PIDNet models.}
    \label{tab:resnet50}
\end{table*}

\begin{table*}[!t]
    \centering
    \begin{tabular}{>{\centering}p{0.135\textwidth}
    >{\centering}p{0.135\textwidth}
    >{\centering}p{0.135\textwidth}
    >{\centering}p{0.135\textwidth}
    >{\centering}p{0.135\textwidth}
    p{0.135\textwidth}}
       \textbf{Input} & \textbf{GT} & \textbf{ResNet-101} & \textbf{PIDNet-S} & \textbf{PIDNet-M} & \hspace{6mm}\textbf{PIDNet-L}  \\
       \includegraphics[width=0.155\textwidth]{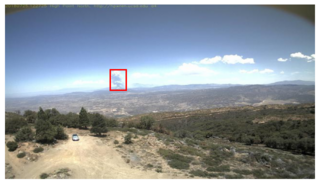} & 
        \includegraphics[width=0.155\textwidth]{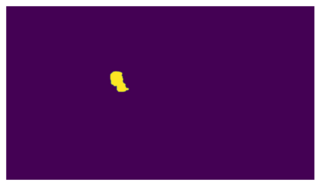} & 
        \includegraphics[width=0.155\textwidth]{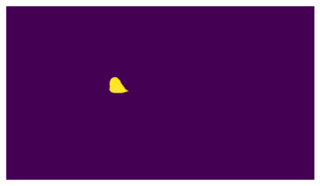} &
        \includegraphics[width=0.155\textwidth]{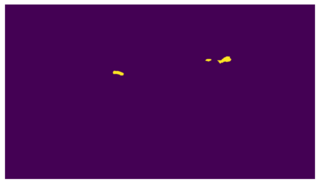} &
        \includegraphics[width=0.155\textwidth]{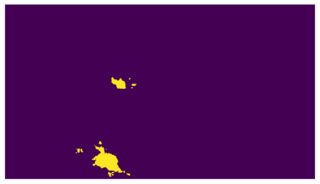} &
        \includegraphics[width=0.155\textwidth]{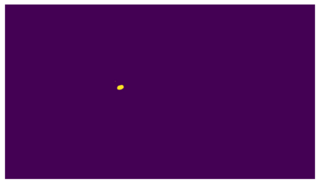} \\

        \includegraphics[width=0.155\textwidth]{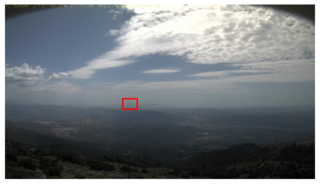} & 
        \includegraphics[width=0.155\textwidth]{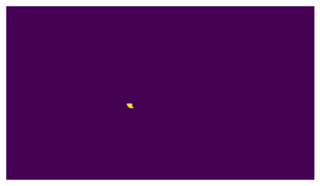} & 
        \includegraphics[width=0.155\textwidth]{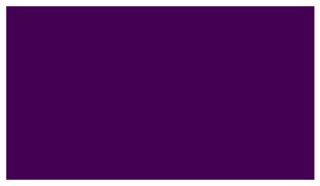} &
        \includegraphics[width=0.155\textwidth]{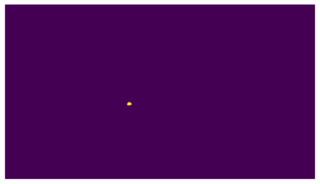} &
        \includegraphics[width=0.155\textwidth]{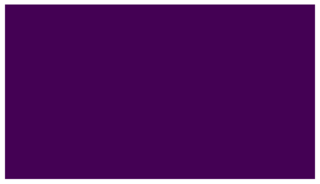} &
        \includegraphics[width=0.155\textwidth]{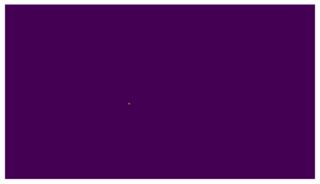} \\

        \includegraphics[width=0.155\textwidth]{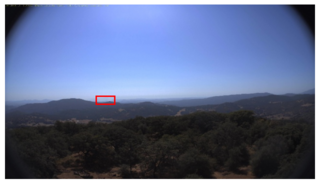} & 
        \includegraphics[width=0.155\textwidth]{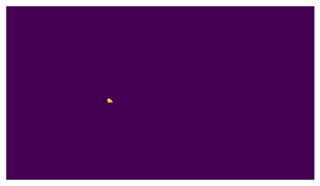} & 
        \includegraphics[width=0.155\textwidth]{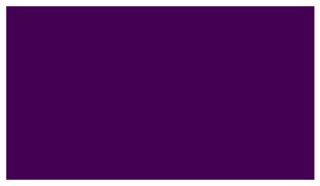} &
        \includegraphics[width=0.155\textwidth]{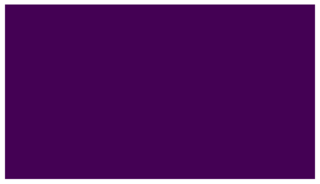} &
        \includegraphics[width=0.155\textwidth]{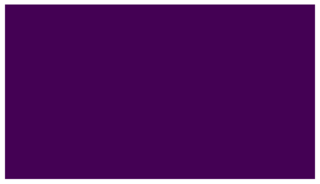} &
        \includegraphics[width=0.155\textwidth]{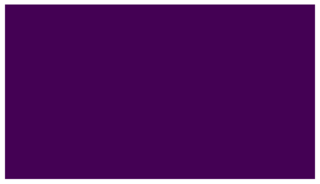} \\

        \includegraphics[width=0.155\textwidth]{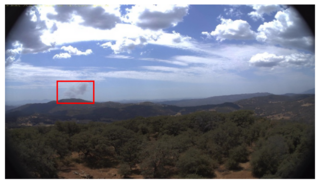} & 
        \includegraphics[width=0.155\textwidth]{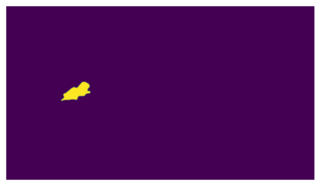} & 
        \includegraphics[width=0.155\textwidth]{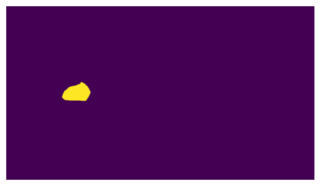} &
        \includegraphics[width=0.155\textwidth]{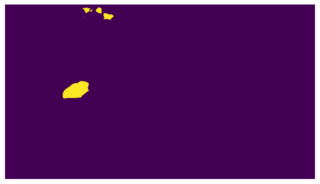} &
        \includegraphics[width=0.155\textwidth]{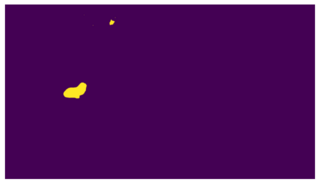} &
        \includegraphics[width=0.155\textwidth]{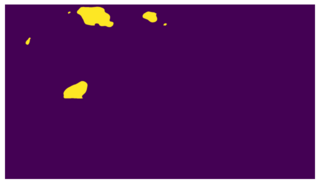} \\

        \includegraphics[width=0.155\textwidth]{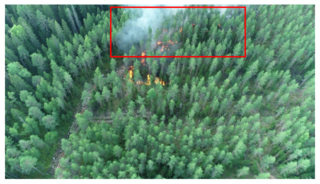} & 
        \includegraphics[width=0.155\textwidth]{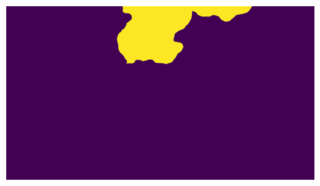} & 
        \includegraphics[width=0.155\textwidth]{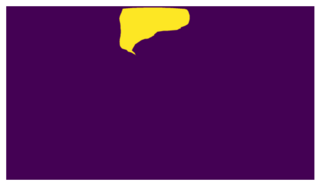} &
        \includegraphics[width=0.155\textwidth]{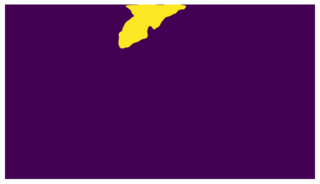} &
        \includegraphics[width=0.155\textwidth]{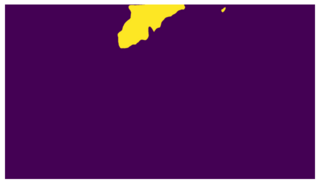} &
        \includegraphics[width=0.155\textwidth]{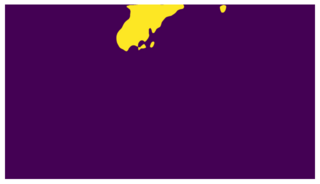} \\

        \includegraphics[width=0.155\textwidth]{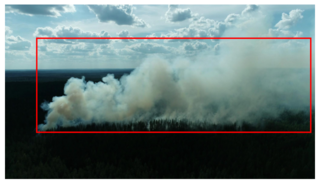} & 
        \includegraphics[width=0.155\textwidth]{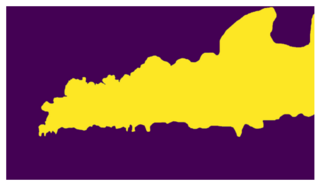} & 
        \includegraphics[width=0.155\textwidth]{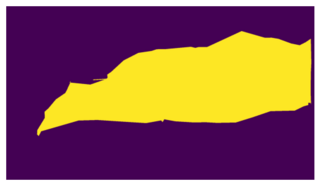} &
        \includegraphics[width=0.155\textwidth]{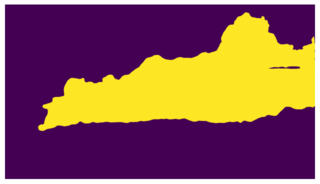} &
        \includegraphics[width=0.155\textwidth]{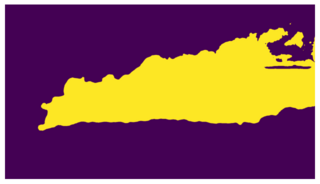} &
        \includegraphics[width=0.155\textwidth]{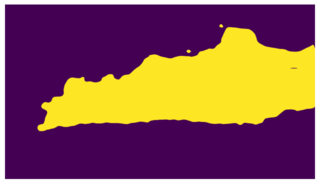} \\

        \includegraphics[width=0.155\textwidth]{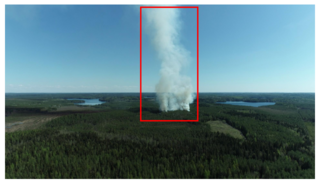} & 
        \includegraphics[width=0.155\textwidth]{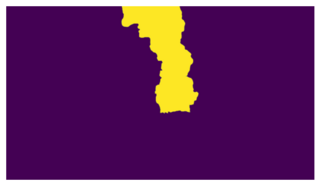} & 
        \includegraphics[width=0.155\textwidth]{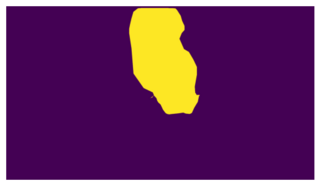} &
        \includegraphics[width=0.155\textwidth]{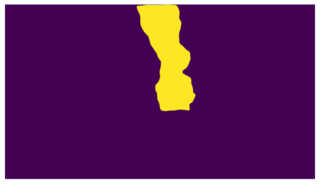} &
        \includegraphics[width=0.155\textwidth]{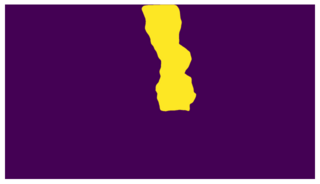} &
        \includegraphics[width=0.155\textwidth]{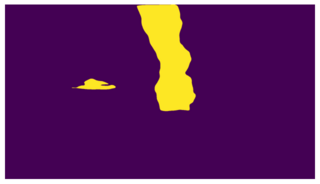} \\

        \includegraphics[width=0.155\textwidth]{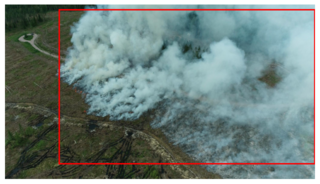} & 
        \includegraphics[width=0.155\textwidth]{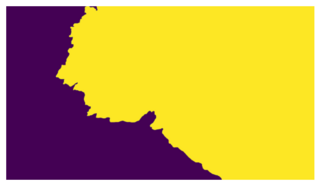} & 
        \includegraphics[width=0.155\textwidth]{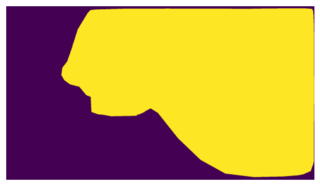} &
        \includegraphics[width=0.155\textwidth]{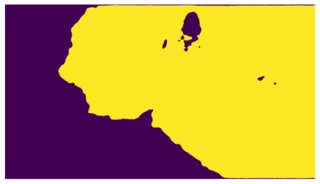} &
        \includegraphics[width=0.155\textwidth]{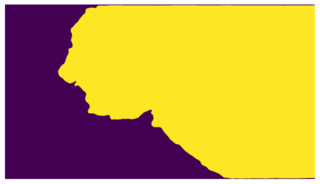} &
        \includegraphics[width=0.155\textwidth]{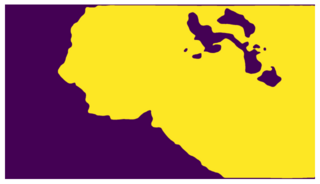} \\
    \end{tabular}
    \caption{Visualisation of outputs obtained using the BoxSnake supervised ResNet-101 and the corresponding distilled PIDNet models.}
    \label{tab:resnet101}
\end{table*}

\begin{table*}[!t]
    \centering
    \begin{tabular}{>{\centering}p{0.135\textwidth}
    >{\centering}p{0.135\textwidth}
    >{\centering}p{0.135\textwidth}
    >{\centering}p{0.135\textwidth}
    >{\centering}p{0.135\textwidth}
    p{0.135\textwidth}}
       \textbf{Input} & \textbf{GT} & \textbf{Swin-B} & \textbf{PIDNet-S} & \textbf{PIDNet-M} & \hspace{6mm}\textbf{PIDNet-L}  \\
       \includegraphics[width=0.155\textwidth]{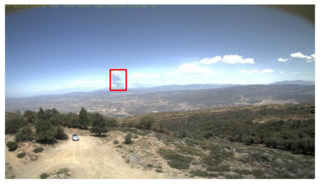} & 
        \includegraphics[width=0.155\textwidth]{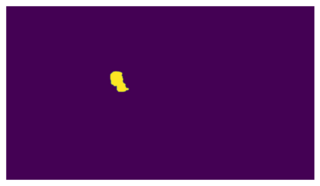} & 
        \includegraphics[width=0.155\textwidth]{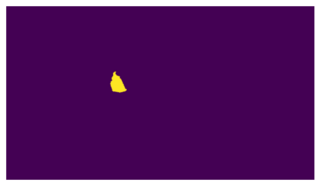} &
        \includegraphics[width=0.155\textwidth]{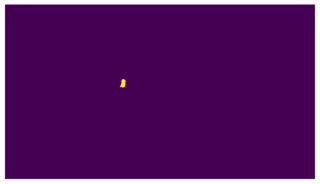} &
        \includegraphics[width=0.155\textwidth]{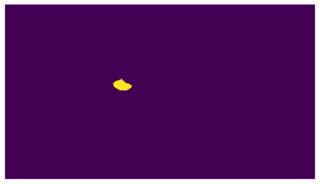} &
        \includegraphics[width=0.155\textwidth]{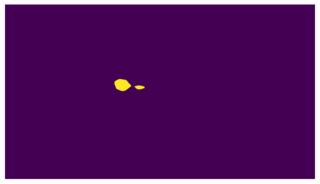} \\

        \includegraphics[width=0.155\textwidth]{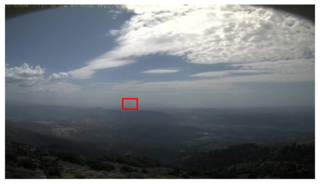} & 
        \includegraphics[width=0.155\textwidth]{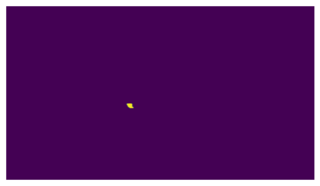} & 
        \includegraphics[width=0.155\textwidth]{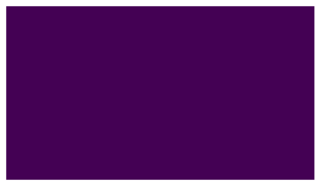} &
        \includegraphics[width=0.155\textwidth]{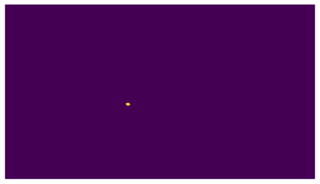} &
        \includegraphics[width=0.155\textwidth]{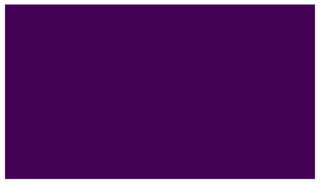} &
        \includegraphics[width=0.155\textwidth]{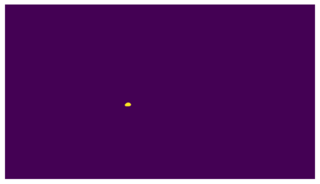} \\

        \includegraphics[width=0.155\textwidth]{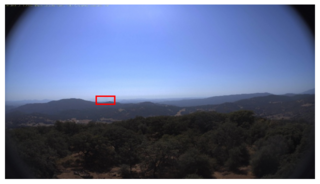} & 
        \includegraphics[width=0.155\textwidth]{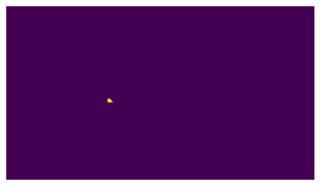} & 
        \includegraphics[width=0.155\textwidth]{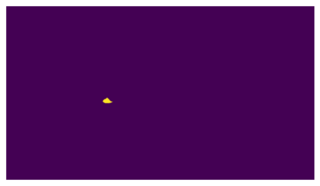} &
        \includegraphics[width=0.155\textwidth]{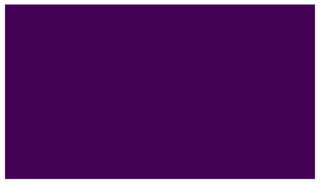} &
        \includegraphics[width=0.155\textwidth]{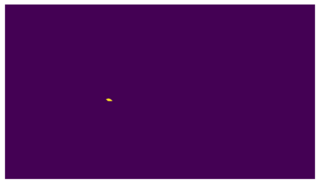} &
        \includegraphics[width=0.155\textwidth]{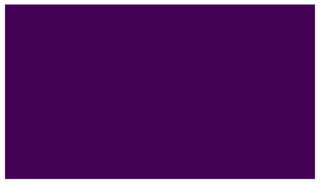} \\

        \includegraphics[width=0.155\textwidth]{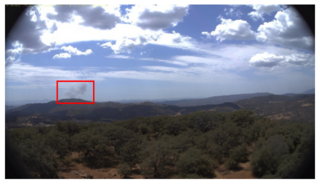} & 
        \includegraphics[width=0.155\textwidth]{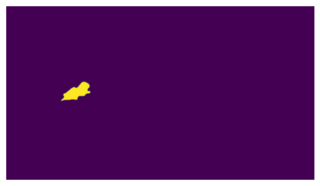} & 
        \includegraphics[width=0.155\textwidth]{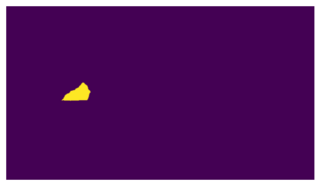} &
        \includegraphics[width=0.155\textwidth]{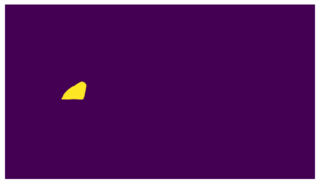} &
        \includegraphics[width=0.155\textwidth]{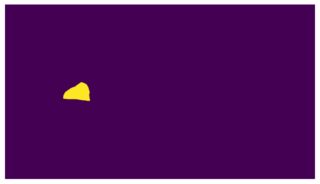} &
        \includegraphics[width=0.155\textwidth]{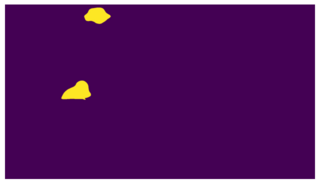} \\

        \includegraphics[width=0.155\textwidth]{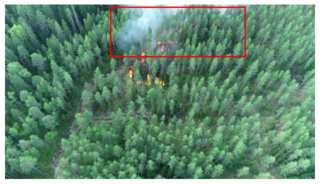} & 
        \includegraphics[width=0.155\textwidth]{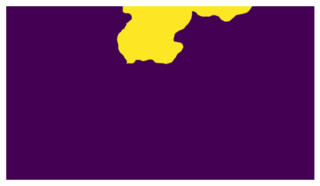} & 
        \includegraphics[width=0.155\textwidth]{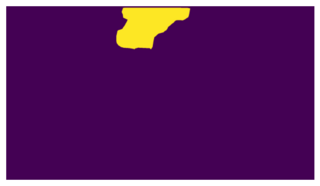} &
        \includegraphics[width=0.155\textwidth]{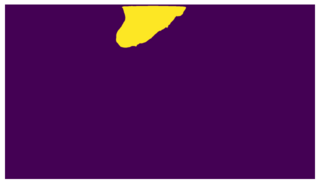} &
        \includegraphics[width=0.155\textwidth]{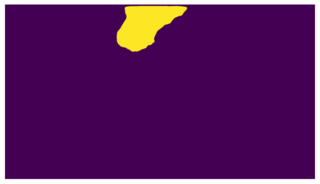} &
        \includegraphics[width=0.155\textwidth]{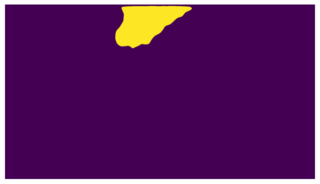} \\

        \includegraphics[width=0.155\textwidth]{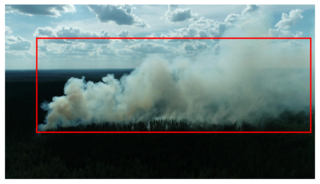} & 
        \includegraphics[width=0.155\textwidth]{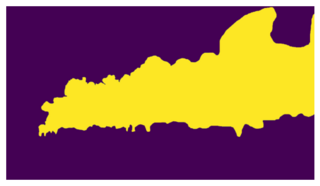} & 
        \includegraphics[width=0.155\textwidth]{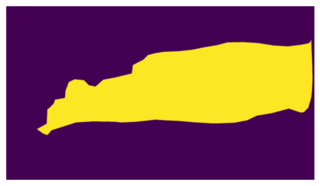} &
        \includegraphics[width=0.155\textwidth]{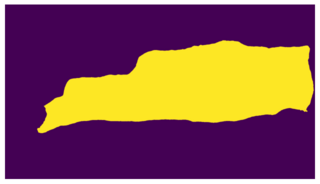} &
        \includegraphics[width=0.155\textwidth]{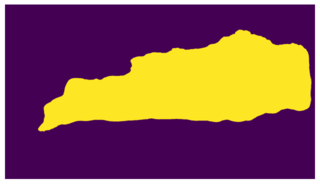} &
        \includegraphics[width=0.155\textwidth]{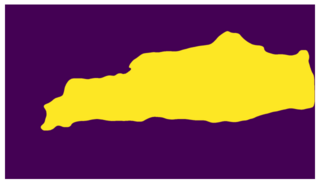} \\

        \includegraphics[width=0.155\textwidth]{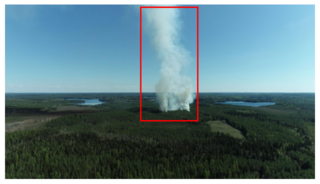} & 
        \includegraphics[width=0.155\textwidth]{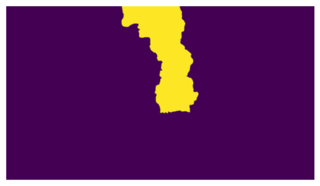} & 
        \includegraphics[width=0.155\textwidth]{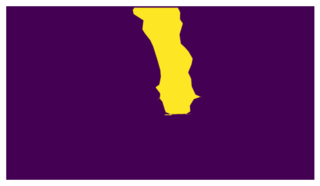} &
        \includegraphics[width=0.155\textwidth]{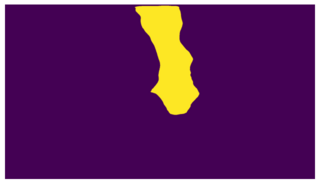} &
        \includegraphics[width=0.155\textwidth]{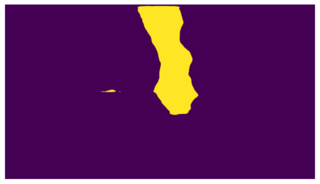} &
        \includegraphics[width=0.155\textwidth]{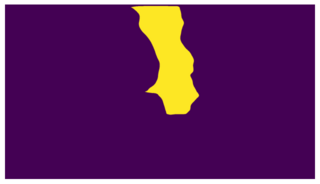} \\

        \includegraphics[width=0.155\textwidth]{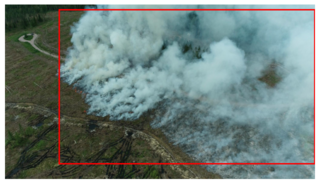} & 
        \includegraphics[width=0.155\textwidth]{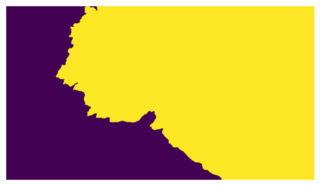} & 
        \includegraphics[width=0.155\textwidth]{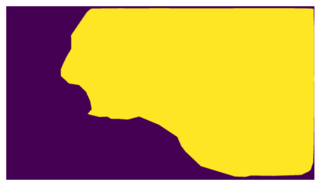} &
        \includegraphics[width=0.155\textwidth]{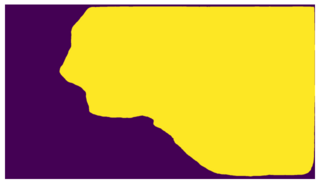} &
        \includegraphics[width=0.155\textwidth]{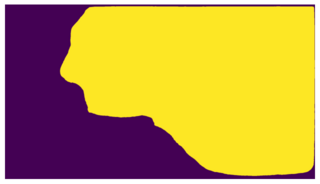} &
        \includegraphics[width=0.155\textwidth]{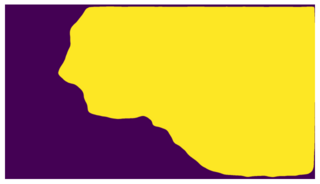} \\
    \end{tabular}
    \caption{Visualisation of outputs obtained using the BoxSnake supervised Swin-B and the corresponding distilled PIDNet models.}
    \label{tab:swin_b}
\end{table*}

\begin{table*}[!t]
    \centering
    \begin{tabular}{>{\centering}p{0.135\textwidth}
    >{\centering}p{0.135\textwidth}
    >{\centering}p{0.135\textwidth}
    >{\centering}p{0.135\textwidth}
    >{\centering}p{0.135\textwidth}
    p{0.135\textwidth}}
       \textbf{Input} & \textbf{GT} & \textbf{Swin-L} & \textbf{PIDNet-S} & \textbf{PIDNet-M} & \hspace{6mm}\textbf{PIDNet-L}  \\
       \includegraphics[width=0.155\textwidth]{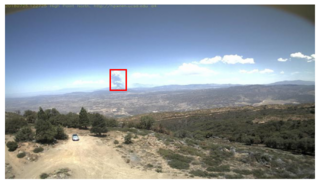} & 
        \includegraphics[width=0.155\textwidth]{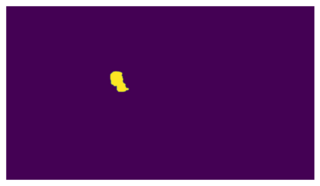} & 
        \includegraphics[width=0.155\textwidth]{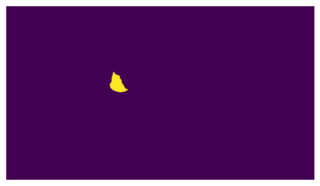} &
        \includegraphics[width=0.155\textwidth]{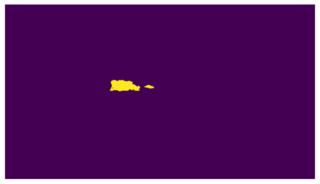} &
        \includegraphics[width=0.155\textwidth]{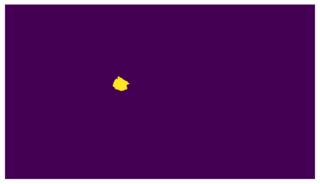} &
        \includegraphics[width=0.155\textwidth]{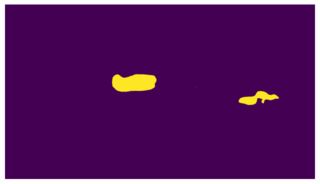} \\

        \includegraphics[width=0.155\textwidth]{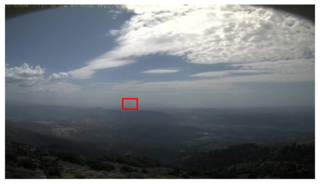} & 
        \includegraphics[width=0.155\textwidth]{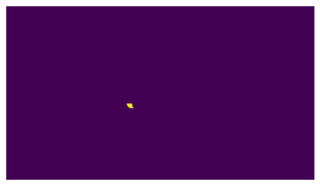} & 
        \includegraphics[width=0.155\textwidth]{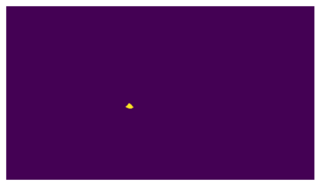} &
        \includegraphics[width=0.155\textwidth]{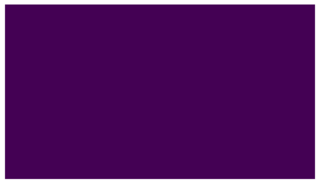} &
        \includegraphics[width=0.155\textwidth]{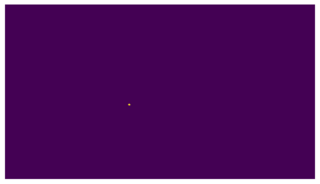} &
        \includegraphics[width=0.155\textwidth]{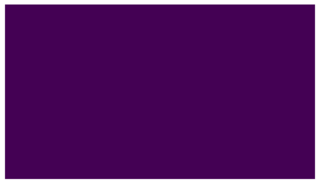} \\

        \includegraphics[width=0.155\textwidth]{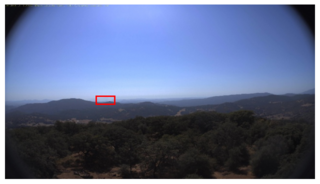} & 
        \includegraphics[width=0.155\textwidth]{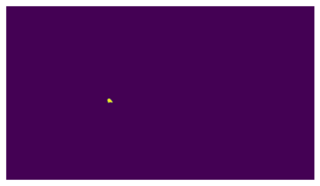} & 
        \includegraphics[width=0.155\textwidth]{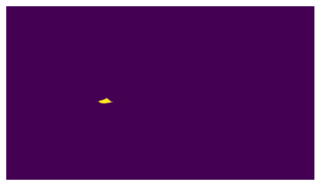} &
        \includegraphics[width=0.155\textwidth]{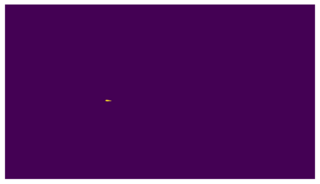} &
        \includegraphics[width=0.155\textwidth]{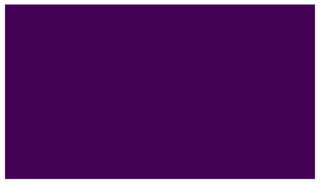} &
        \includegraphics[width=0.155\textwidth]{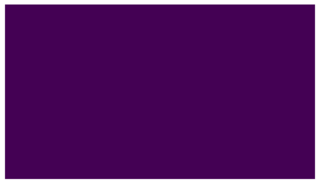} \\

        \includegraphics[width=0.155\textwidth]{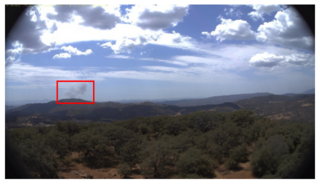} & 
        \includegraphics[width=0.155\textwidth]{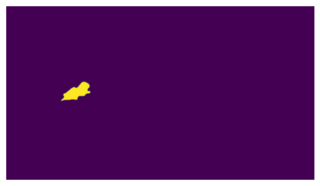} & 
        \includegraphics[width=0.155\textwidth]{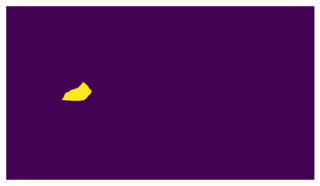} &
        \includegraphics[width=0.155\textwidth]{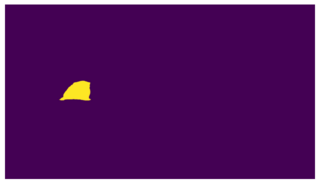} &
        \includegraphics[width=0.155\textwidth]{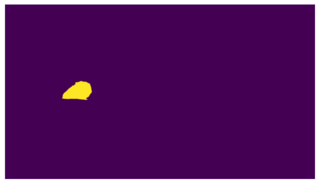} &
        \includegraphics[width=0.155\textwidth]{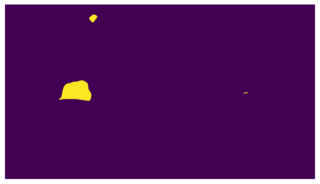} \\

        \includegraphics[width=0.155\textwidth]{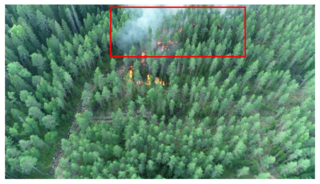} & 
        \includegraphics[width=0.155\textwidth]{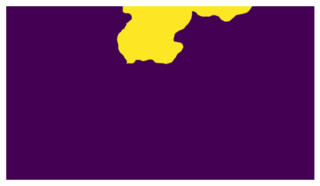} & 
        \includegraphics[width=0.155\textwidth]{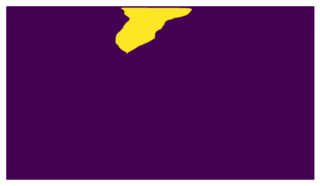} &
        \includegraphics[width=0.155\textwidth]{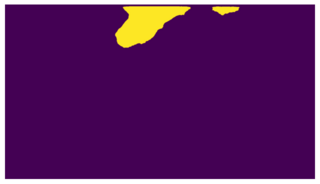} &
        \includegraphics[width=0.155\textwidth]{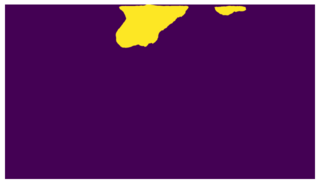} &
        \includegraphics[width=0.155\textwidth]{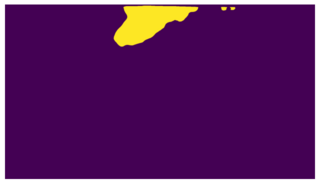} \\

        \includegraphics[width=0.155\textwidth]{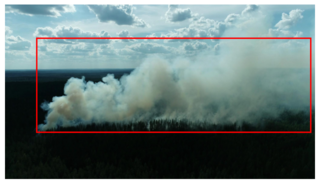} & 
        \includegraphics[width=0.155\textwidth]{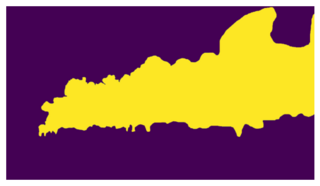} & 
        \includegraphics[width=0.155\textwidth]{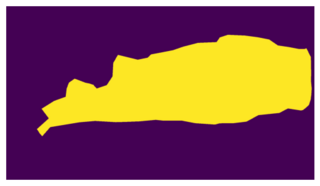} &
        \includegraphics[width=0.155\textwidth]{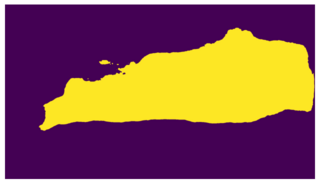} &
        \includegraphics[width=0.155\textwidth]{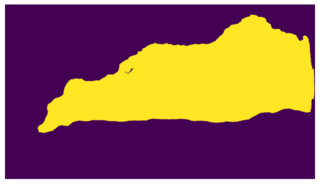} &
        \includegraphics[width=0.155\textwidth]{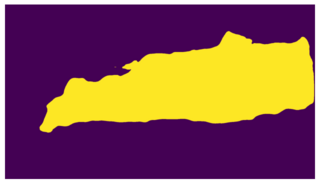} \\

        \includegraphics[width=0.155\textwidth]{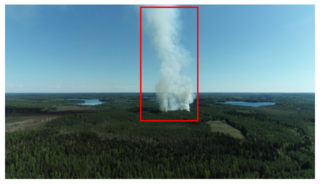} & 
        \includegraphics[width=0.155\textwidth]{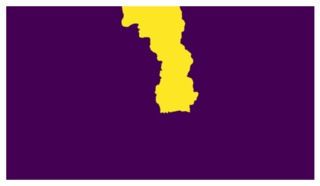} & 
        \includegraphics[width=0.155\textwidth]{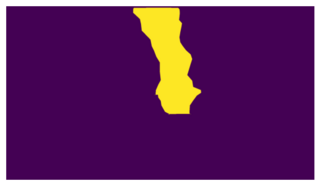} &
        \includegraphics[width=0.155\textwidth]{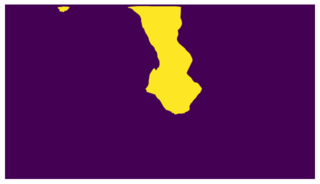} &
        \includegraphics[width=0.155\textwidth]{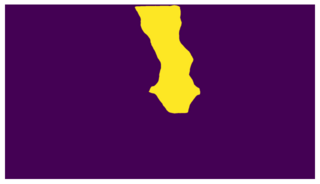} &
        \includegraphics[width=0.155\textwidth]{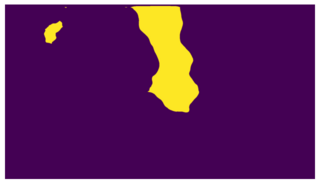} \\

        \includegraphics[width=0.155\textwidth]{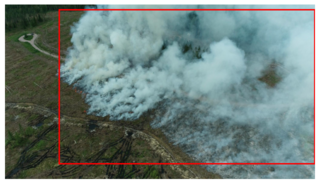} & 
        \includegraphics[width=0.155\textwidth]{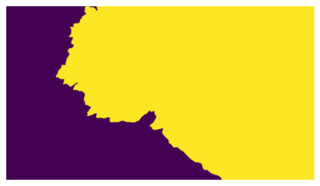} & 
        \includegraphics[width=0.155\textwidth]{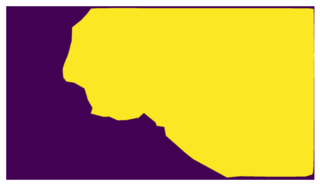} &
        \includegraphics[width=0.155\textwidth]{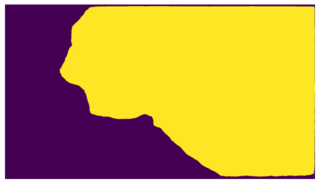} &
        \includegraphics[width=0.155\textwidth]{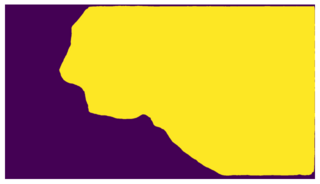} &
        \includegraphics[width=0.155\textwidth]{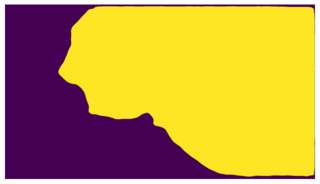} \\
    \end{tabular}
    \caption{Visualisation of outputs obtained using the BoxSnake supervised Swin-L and the corresponding distilled PIDNet models.}
    \label{tab:swin_l}
\end{table*}

\begin{table*}[!t]
    \centering
    \begin{tabular}{>{\centering}p{0.135\textwidth}
    >{\centering}p{0.135\textwidth}
    >{\centering}p{0.135\textwidth}
    >{\centering}p{0.135\textwidth}
    >{\centering}p{0.135\textwidth}
    p{0.135\textwidth}}
       \textbf{Input} & \textbf{GT} & \textbf{SAM} & \textbf{PIDNet-S} & \textbf{PIDNet-M} & \hspace{6mm}\textbf{PIDNet-L}  \\
       \includegraphics[width=0.155\textwidth]{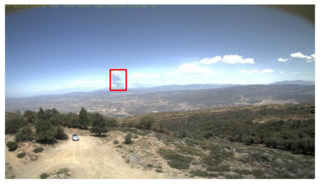} & 
        \includegraphics[width=0.155\textwidth]{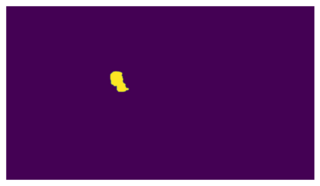} & 
        \includegraphics[width=0.155\textwidth]{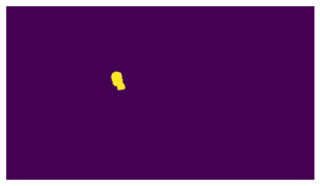} &
        \includegraphics[width=0.155\textwidth]{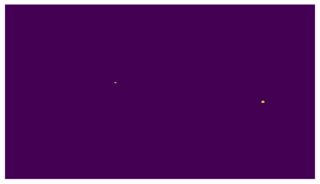} &
        \includegraphics[width=0.155\textwidth]{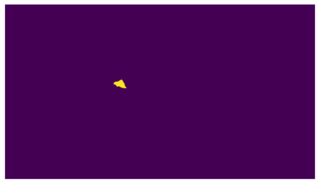} &
        \includegraphics[width=0.155\textwidth]{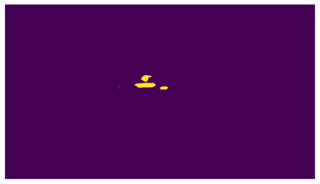} \\

        \includegraphics[width=0.155\textwidth]{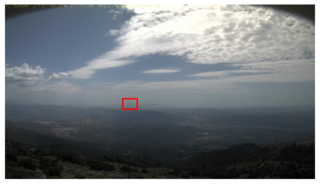} & 
        \includegraphics[width=0.155\textwidth]{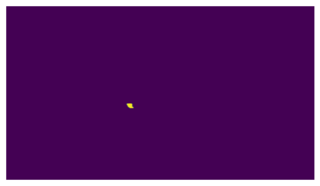} & 
        \includegraphics[width=0.155\textwidth]{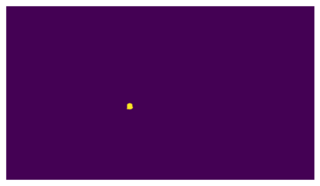} &
        \includegraphics[width=0.155\textwidth]{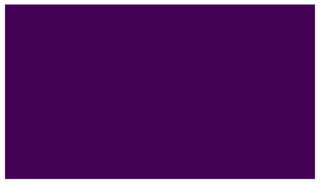} &
        \includegraphics[width=0.155\textwidth]{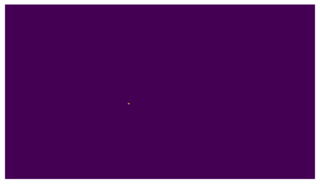} &
        \includegraphics[width=0.155\textwidth]{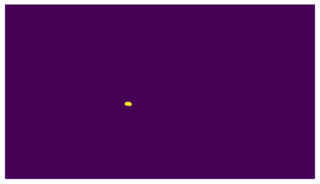} \\

        \includegraphics[width=0.155\textwidth]{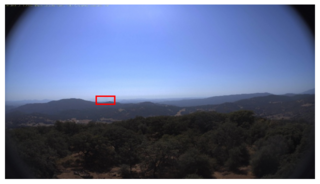} & 
        \includegraphics[width=0.155\textwidth]{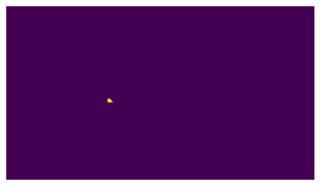} & 
        \includegraphics[width=0.155\textwidth]{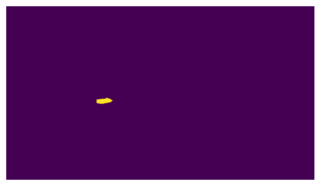} &
        \includegraphics[width=0.155\textwidth]{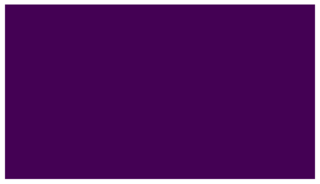} &
        \includegraphics[width=0.155\textwidth]{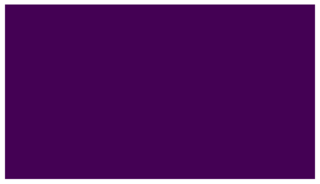} &
        \includegraphics[width=0.155\textwidth]{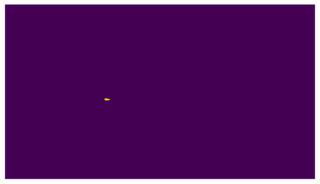} \\

        \includegraphics[width=0.155\textwidth]{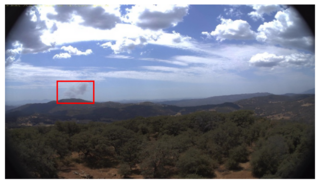} & 
        \includegraphics[width=0.155\textwidth]{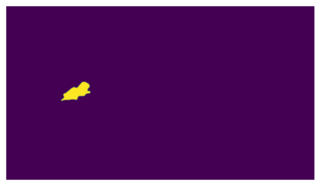} & 
        \includegraphics[width=0.155\textwidth]{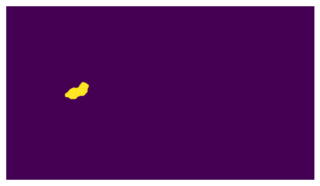} &
        \includegraphics[width=0.155\textwidth]{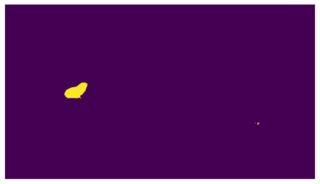} &
        \includegraphics[width=0.155\textwidth]{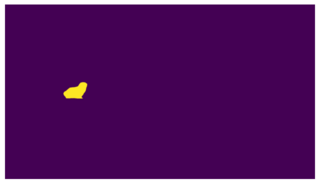} &
        \includegraphics[width=0.155\textwidth]{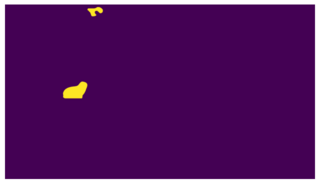} \\

        \includegraphics[width=0.155\textwidth]{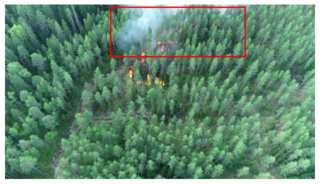} & 
        \includegraphics[width=0.155\textwidth]{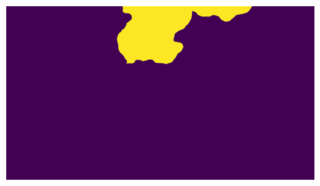} & 
        \includegraphics[width=0.155\textwidth]{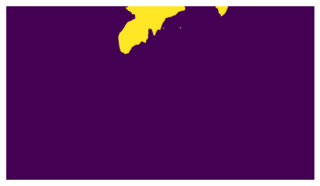} &
        \includegraphics[width=0.155\textwidth]{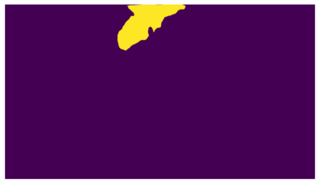} &
        \includegraphics[width=0.155\textwidth]{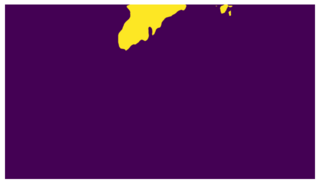} &
        \includegraphics[width=0.155\textwidth]{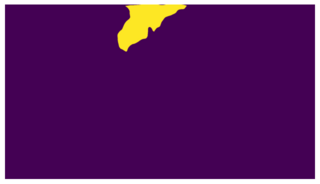} \\

        \includegraphics[width=0.155\textwidth]{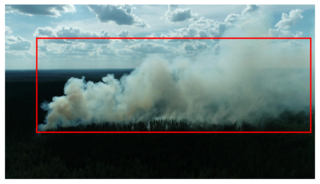} & 
        \includegraphics[width=0.155\textwidth]{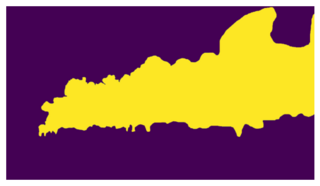} & 
        \includegraphics[width=0.155\textwidth]{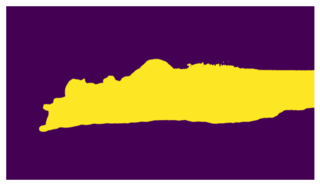} &
        \includegraphics[width=0.155\textwidth]{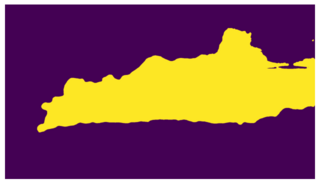} &
        \includegraphics[width=0.155\textwidth]{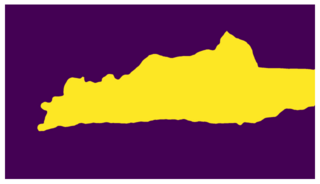} &
        \includegraphics[width=0.155\textwidth]{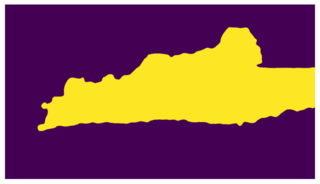} \\

        \includegraphics[width=0.155\textwidth]{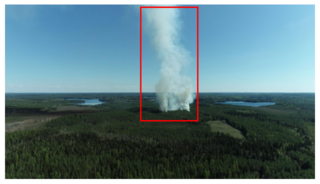} & 
        \includegraphics[width=0.155\textwidth]{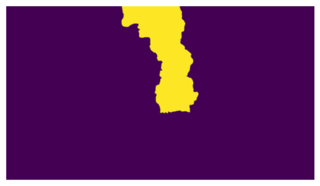} & 
        \includegraphics[width=0.155\textwidth]{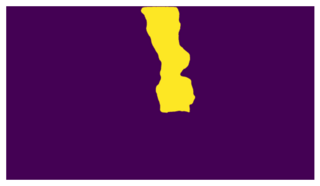} &
        \includegraphics[width=0.155\textwidth]{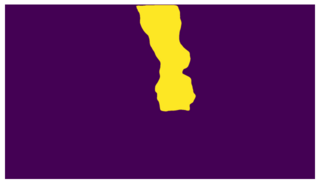} &
        \includegraphics[width=0.155\textwidth]{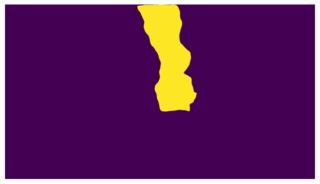} &
        \includegraphics[width=0.155\textwidth]{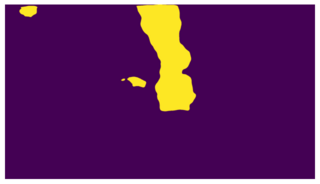} \\

        \includegraphics[width=0.155\textwidth]{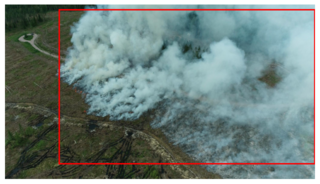} & 
        \includegraphics[width=0.155\textwidth]{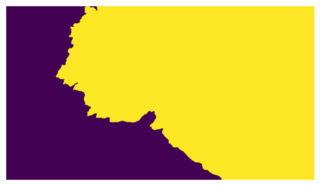} & 
        \includegraphics[width=0.155\textwidth]{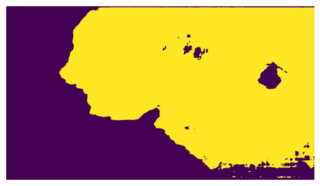} &
        \includegraphics[width=0.155\textwidth]{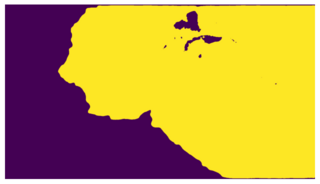} &
        \includegraphics[width=0.155\textwidth]{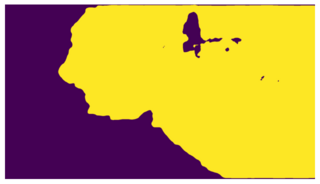} &
        \includegraphics[width=0.155\textwidth]{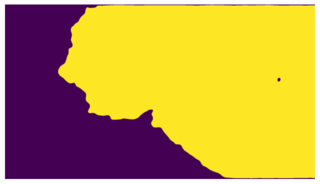} \\
    \end{tabular}
    \caption{Visualisation of outputs obtained using SAM and the corresponding distilled PIDNet models.}
    \label{tab:sam}
\end{table*}

\begin{table*}[!ht]
    \centering
    \begin{tabularx}{\textwidth}{l|rrrrrrrrrrrr} 
        \textbf{Device} & \multicolumn{4}{c|}{CPU} & \multicolumn{8}{c}{GPU} \\
        \hline
        \textbf{TensorRT} & \multicolumn{4}{c|}{-} & \multicolumn{4}{c|}{No} & \multicolumn{4}{c}{Yes} \\
        \hline
        \textbf{Power setting} & 10W & 15W & 25W & \multicolumn{1}{c|}{MAX} & 10W & 15W & 25W & \multicolumn{1}{c|}{MAX} & 10W & 15W & 25W & MAX \\
        \hline
        \textbf{Model} & \multicolumn{12}{c}{\textbf{FPS}} \\
        \hline
        PIDNet-S & 1.22 & 1.40 & 1.32 & 1.71 & 6.31 & 6.25 & 6.19 & 12.47 & 10.90 & 10.77 & 11.97 & 25.88 \\ 
        PIDNet-M & 0.47 & 0.54 & 0.70 & 0.88 & 2.89 & 2.79 & 2.83 & 5.44 & 4.65 & 4.39 & 5.25 & 9.86 \\ 
        PIDNet-L & 0.32 & 0.37 & 0.53 & 0.63 & 2.17 & 2.18 & 2.17 & 4.29 & 3.35 & 3.37 & 3.96 & 7.47 \\ 
    \end{tabularx}
    \caption{PIDNet model frame rate benchmark on the NVIDIA Jetson Orin NX computer with different power supply limitation, GPU, CPU and TensorRT options. With the GPU the inputs and outputs were 1080 $\times$ 1920 pixels and all operations were computed in full precision (32-bit float). For the CPU, images with a resolution of 576 $\times$ 1024 were used. With the 1080 $\times$ 1920 resolution the fps was always below 1. The specifications of the different power settings are shown in Table~\ref{tab:jetson_power_specs}.}
    \label{tab:hw_bench}
\end{table*}

\begin{table*}
\begin{tabularx}{\textwidth}{l|>{\raggedleft\arraybackslash}X>{\raggedleft\arraybackslash}X>{\raggedleft\arraybackslash}X>{\raggedleft\arraybackslash}X}
    \textbf{Power Budget} 	& \textbf{MAXN }&	\textbf{10W} &	\textbf{15W} &	\textbf{25W} \\
    \hline
Online CPU 	& 8 	& 4 	& 4 	& 8 \\
CPU Max Frequency (MHz) 	& 1984 	& 1190.4 	& 1420.8 	& 1497.6\\
GPU Max Frecuency (MHz) 	& 918 	& 612 	& 612 	& 408\\
DLA cores & 2 & 1 & 1 & 2 \\
DLA Core Max Frequency (MHz) 	& 614.4 	& 153.6 	& 614.4 	& 614.4\\
DLA Falcon Max Frequency (MHz) & 614.4 & 153.6  & 614.4 & 614.4 \\
PVA Cores 	& 1 	& 0 	& 0 	& 1 \\
PVA VPS Max Frequency (MHz) 	& 704 	& N/A 	& N/A 	& 512 \\
PVA AXI Max Frequency (MHz) 	& 486.4 	& N/A 	& N/A 	& 358.4 \\
\end{tabularx}
\caption{The specifications of the different NVIDIA Jetson Orin NX power settings~\cite{JetsonPower}.}
\label{tab:jetson_power_specs}
\end{table*}


\begin{table*}
    \centering
    \begin{tabularx}{\textwidth}{ 
   >{\raggedright\arraybackslash}p{4cm}
   >{\raggedright\arraybackslash}p{2cm} 
   >{\raggedleft\arraybackslash}X
   >{\raggedleft\arraybackslash}X
   >{\raggedleft\arraybackslash}X
   >{\raggedleft\arraybackslash}X
    >{\raggedleft\arraybackslash}X
   }
        \textbf{Teacher} & \textbf{Student} & \textbf{mIoU} & \textbf{Accuracy} & \textbf{Precision} & \textbf{Recall}  &  $\boldsymbol{F_1}$ \\
        \hline
        \rowcolor{LightCyan}
        ResNet-50-RCNN-FPN & PIDNet-S & 0.444 & 0.855 & 0.690 & 0.585 & 0.561 \\
        \rowcolor{LightCyan}
        ResNet-101-RCNN-FPN & PIDNet-S & 0.437 & 0.857 & \textbf{0.841} & 0.499 & 0.572 \\
        \rowcolor{LightCyan}
        Swin-B-FPN & PIDNet-S & \underline{0.472} & \textbf{0.874} & 0.735 & \underline{0.600} & \underline{0.585} \\
        \rowcolor{LightCyan}
        Swin-L-FPN & PIDNet-S & \textbf{0.482} & \underline{0.873} & 0.749 & \textbf{0.629} & \textbf{0.607} \\
        \rowcolor{LightCyan}
        SAM & PIDNet-S & 0.410 & 0.848 & \underline{0.824} & 0.481 & 0.534 \\
        ResNet-50-RCNN-FPN & PIDNet-M & \textbf{0.460} & \textbf{0.867 }& 0.750 & \textbf{0.601} & \textbf{0.586} \\
        ResNet-101-RCNN-FPN & PIDNet-M & 0.452 & 0.855 & 0.774 & \underline{0.542} & 0.575 \\
        Swin-B-FPN & PIDNet-M & 0.406 & 0.857 & 0.737 & 0.512 & 0.525 \\
        Swin-L-FPN & PIDNet-M & \underline{0.454} & \textbf{0.867} & \underline{0.822} & 0.509 & \underline{0.582} \\
        SAM & PIDNet-M & 0.374 & 0.832 & \textbf{0.867} & 0.406 & 0.502 \\
        
        \rowcolor{LightCyan}
        ResNet-50-RCNN-FPN & PIDNet-L & \textbf{0.513} & \textbf{0.882} & \underline{0.807} & \underline{0.588} & \textbf{0.637} \\
        \rowcolor{LightCyan}
        ResNet-101-RCNN-FPN & PIDNet-L & 0.410 & 0.845 & 0.799 & 0.486 & 0.538 \\
        \rowcolor{LightCyan}
        Swin-B-FPN & PIDNet-L & \underline{0.507} & \underline{0.881} & 0.796 & 0.593 & \underline{0.633} \\
        \rowcolor{LightCyan}
        Swin-L-FPN & PIDNet-L & 0.463 & 0.862 & 0.696 & \textbf{0.630} & 0.584 \\
        \rowcolor{LightCyan}
        SAM & PIDNet-L & 0.479 & 0.873 & \textbf{0.823} & 0.552 & 0.614 \\
    \end{tabularx}
    \caption{Test set metrics of the different student models on the Croatian Center for Wildfire Research data. In all metrics higher is better. All teacher models besides SAM were trained using BoxSnake. The best results for each student model are in bold and the second best are underlined.}
    \label{tab:croatia_fullset}
\end{table*}

\begin{table*}[!t]
    \centering
    \begin{tabular}{>{\centering}p{0.17\textwidth}
    >{\centering}p{0.17\textwidth}
    >{\centering}p{0.17\textwidth}
    >{\centering}p{0.17\textwidth}p{0.17\textwidth}}
       \textbf{Input} & \textbf{GT} & \textbf{PIDNet-S} & \textbf{PIDNet-M} & \hspace{6mm}\textbf{PIDNet-L}  \\
       \includegraphics[width=0.18\textwidth]{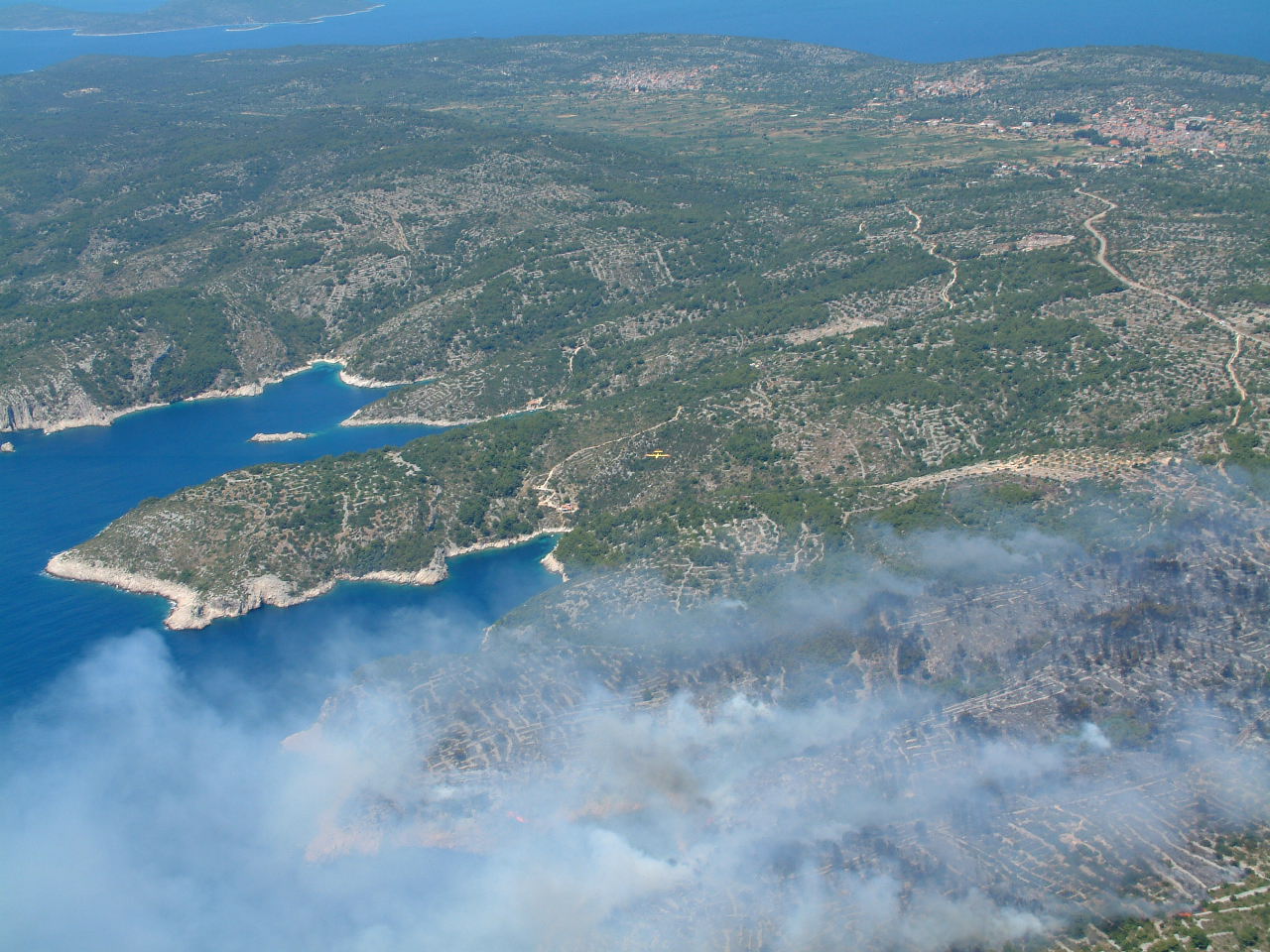} & 
        \includegraphics[width=0.18\textwidth]{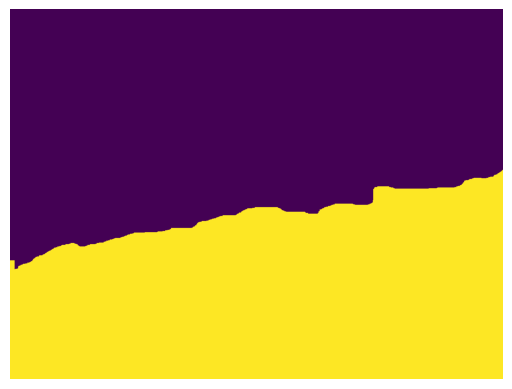} & 
        \includegraphics[width=0.18\textwidth]{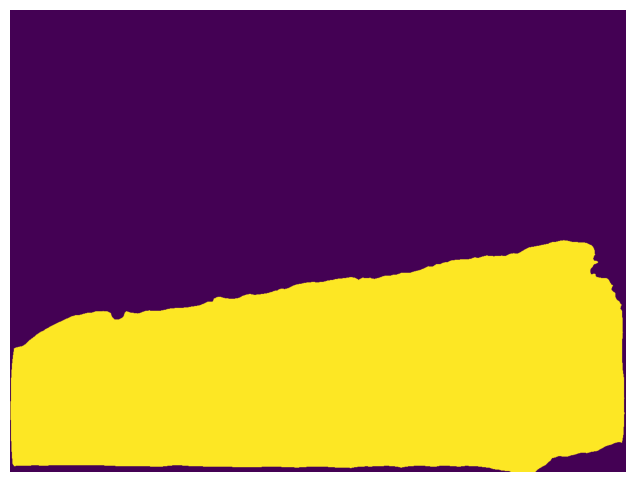} &
        \includegraphics[width=0.18\textwidth]{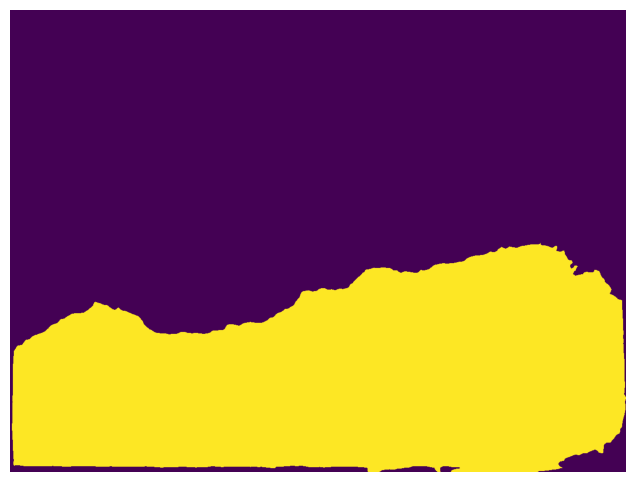} &
        \includegraphics[width=0.18\textwidth]{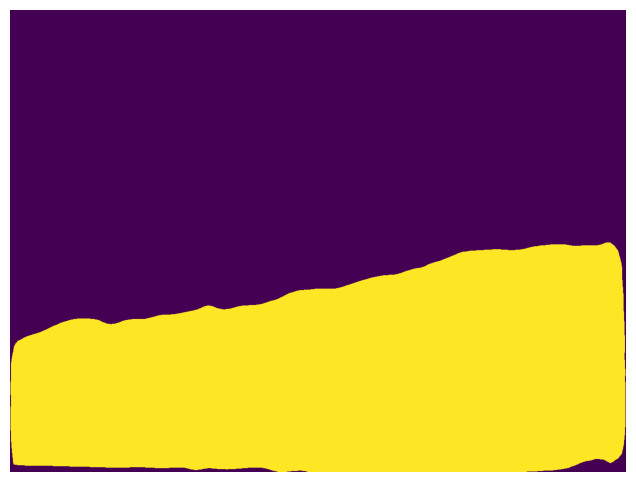} \\
        
        \includegraphics[width=0.18\textwidth]{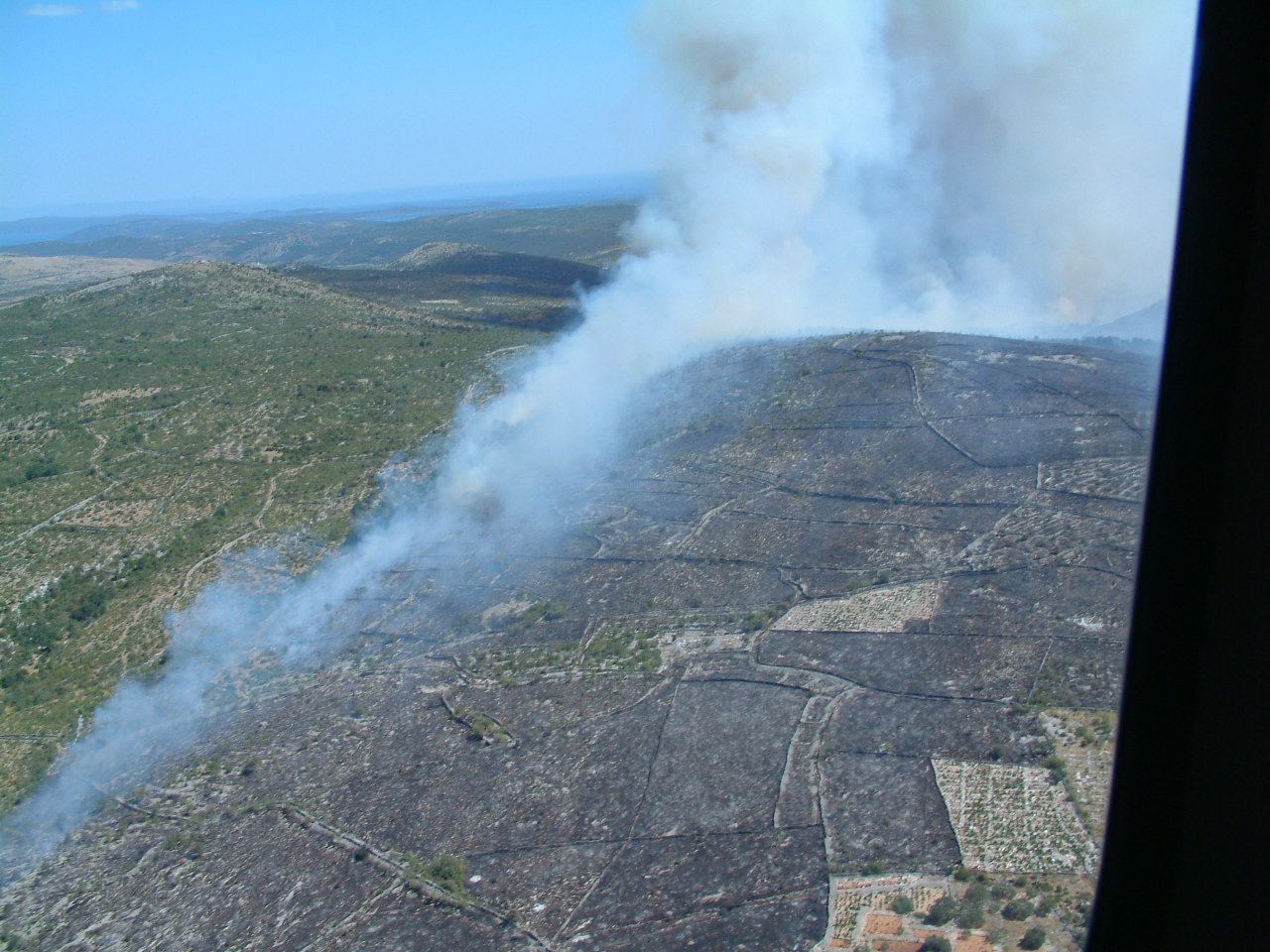} & 
        \includegraphics[width=0.18\textwidth]{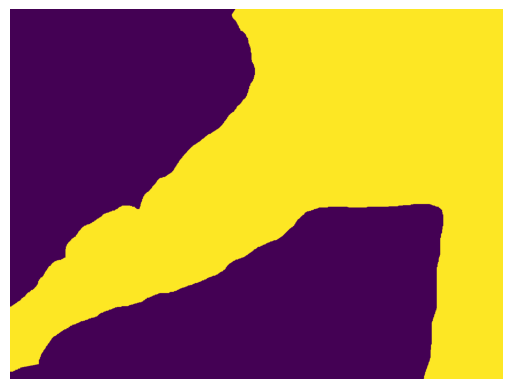} & 
        \includegraphics[width=0.18\textwidth]{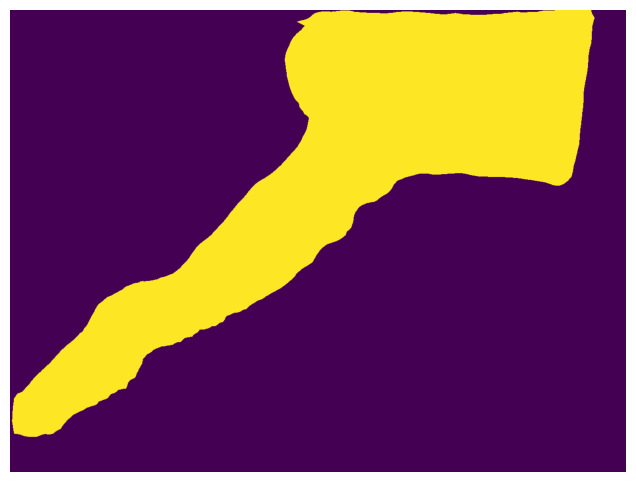} &
        \includegraphics[width=0.18\textwidth]{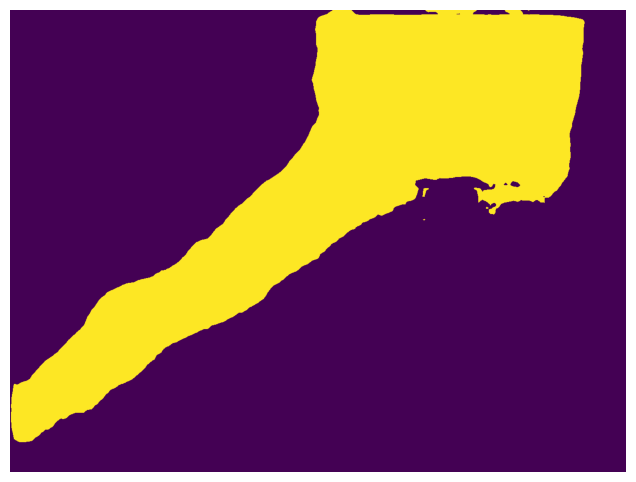} &
        \includegraphics[width=0.18\textwidth]{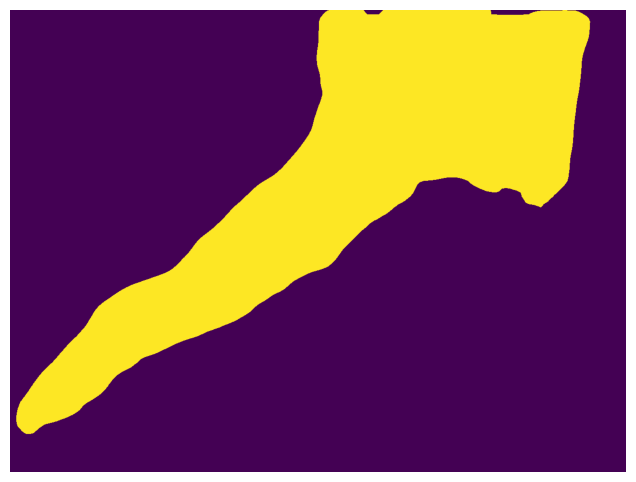} \\

        \includegraphics[width=0.18\textwidth]{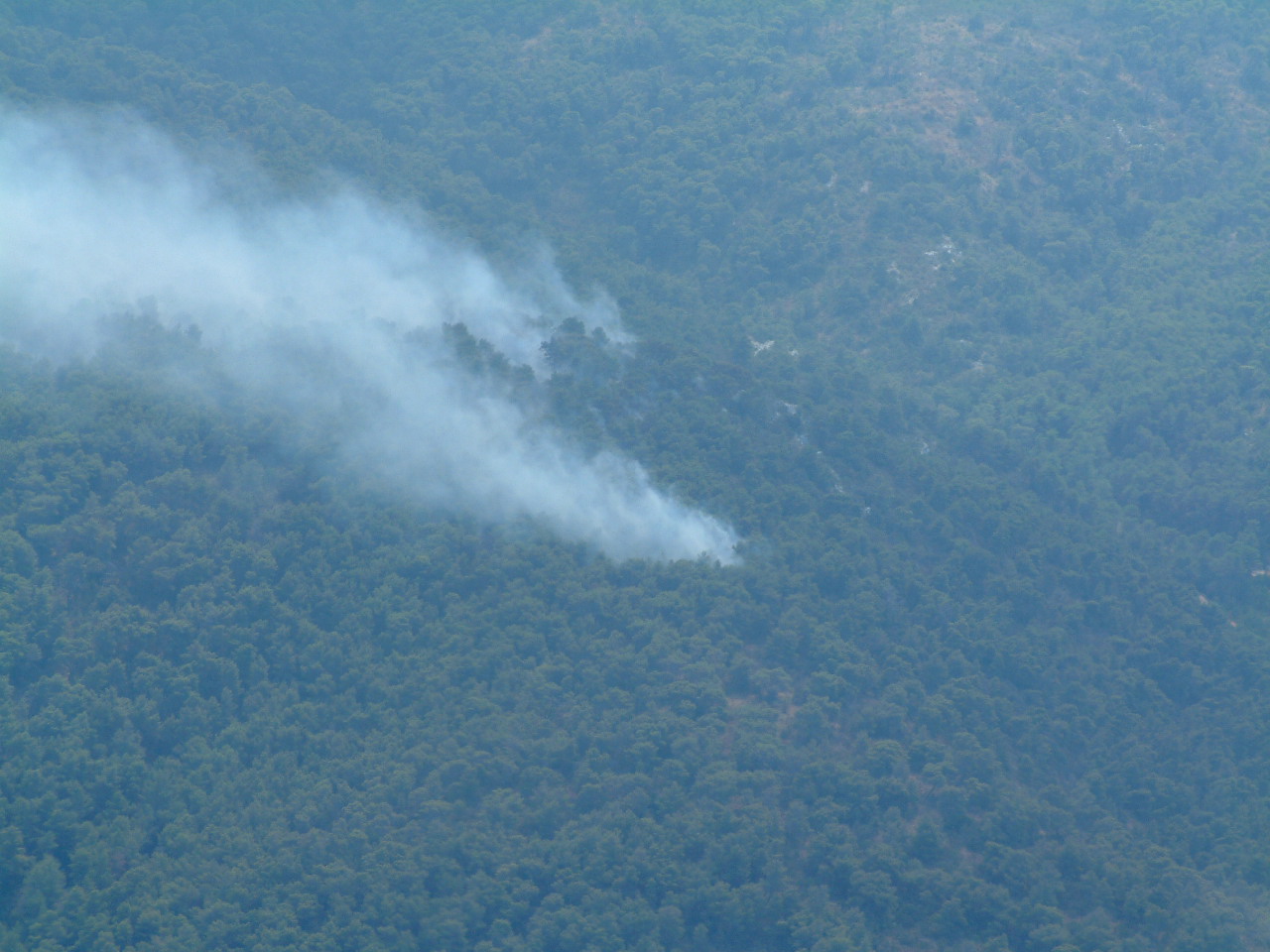} & 
        \includegraphics[width=0.18\textwidth]{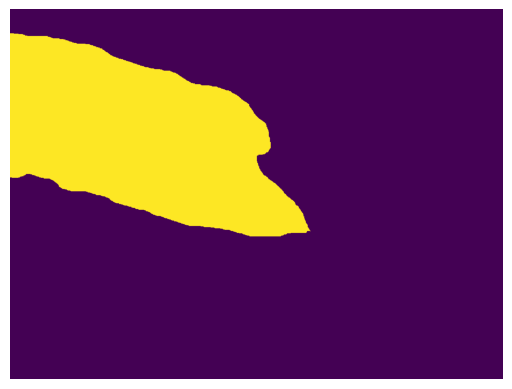} & 
        \includegraphics[width=0.18\textwidth]{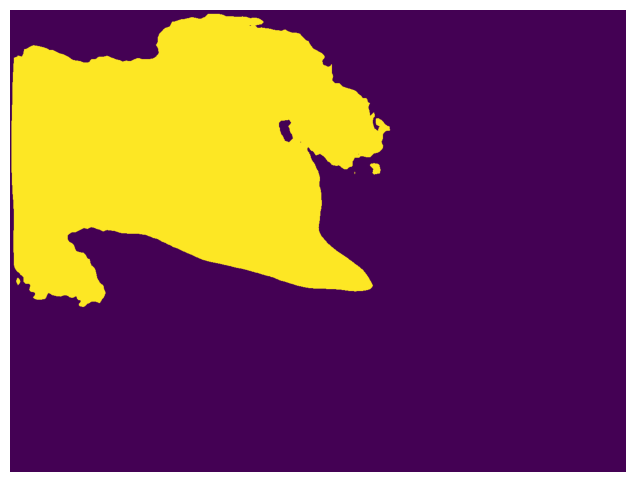} &
        \includegraphics[width=0.18\textwidth]{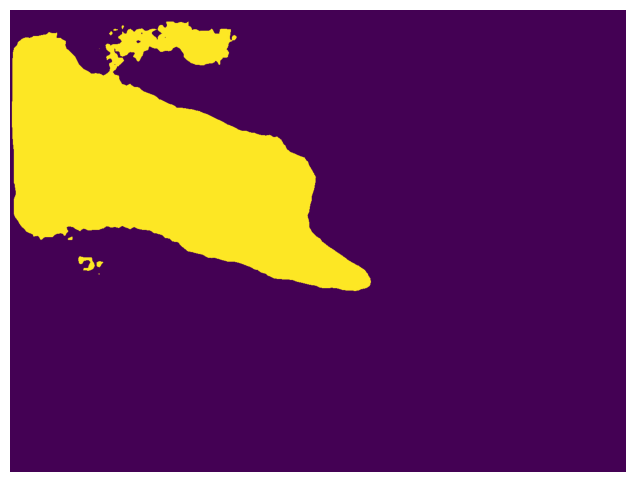} &
        \includegraphics[width=0.18\textwidth]{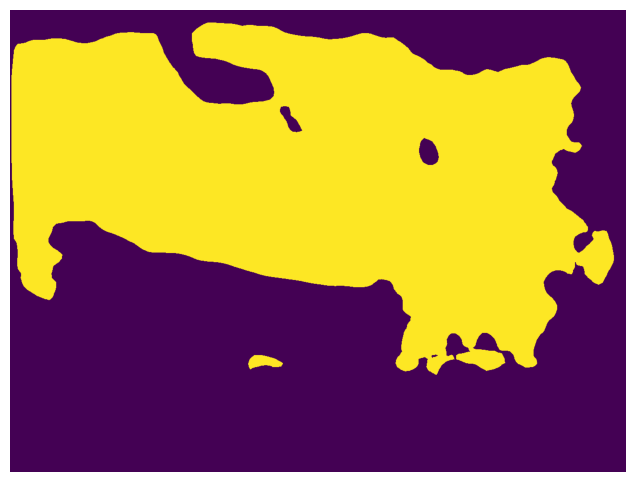} \\

        \includegraphics[width=0.18\textwidth]{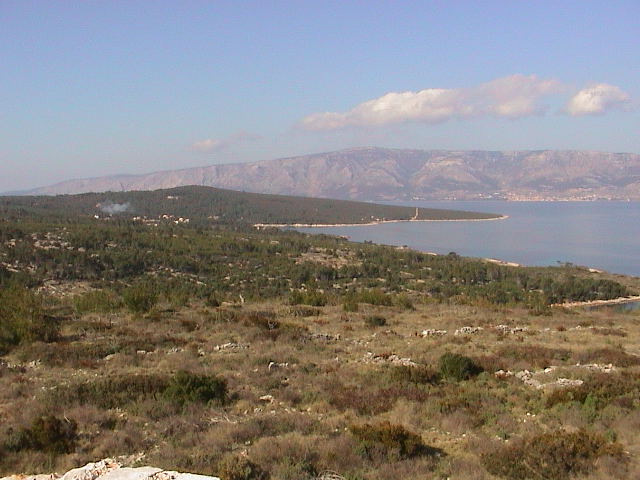} & 
        \includegraphics[width=0.18\textwidth]{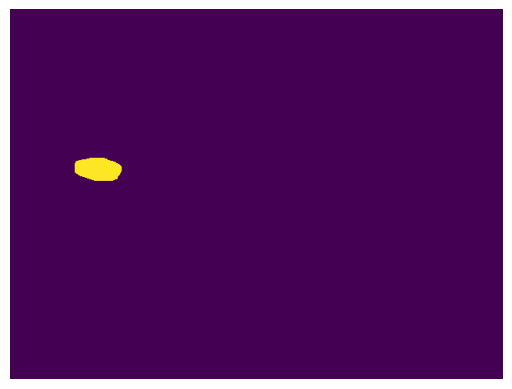} & 
        \includegraphics[width=0.18\textwidth]{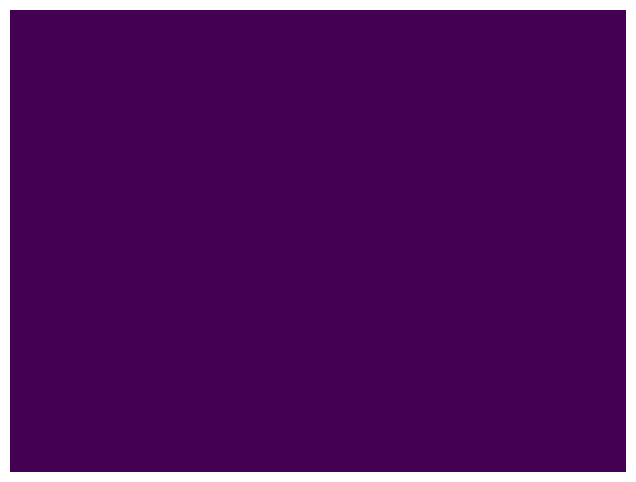} &
        \includegraphics[width=0.18\textwidth]{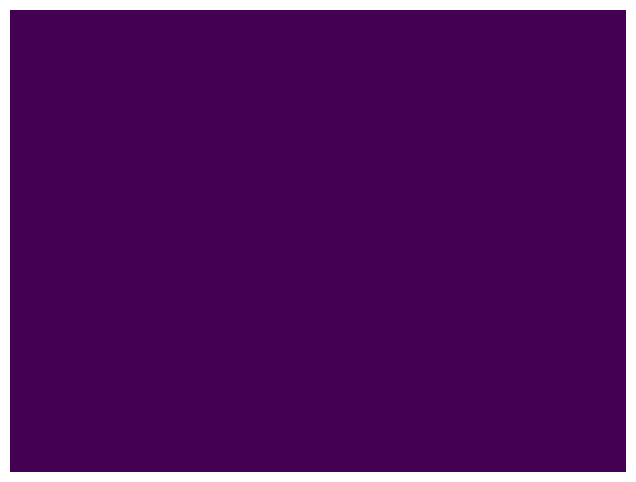} &
        \includegraphics[width=0.18\textwidth]{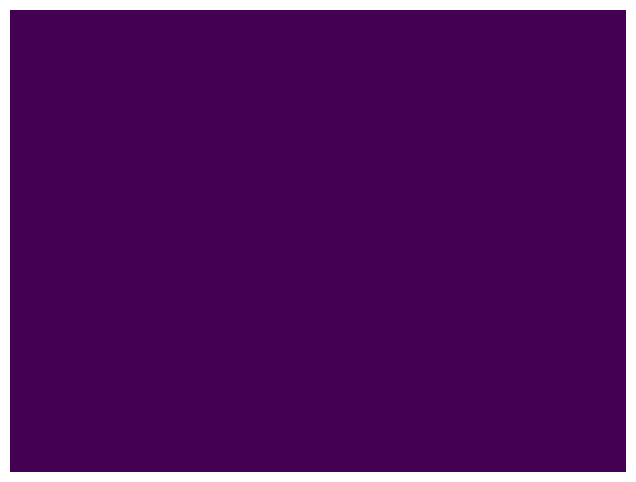} \\
    \end{tabular}
    \caption{Qualitative results of smoke predictions obtained on the Croatian Center for Wildfire Research data using the different SAM supervised models.}
    \label{tab:croatia_qual}
\end{table*}

\end{document}